\newif\ifMain
\newif\ifSup
\Maintrue 
\Suptrue 

\documentclass[10pt,twocolumn,letterpaper]{article}

\usepackage[pagenumbers]{cvpr} 
\usepackage{gensymb}
\usepackage{array}
\usepackage{pifont}
\definecolor{rowgray}{gray}{0.95}
\newcommand{\cmark}{\ding{51}} 
\newcommand{\xmark}{\ding{55}} 
\usepackage{graphicx}
\usepackage{amsmath}
\usepackage{multirow}
\usepackage{booktabs}
\usepackage{array}
\usepackage{float} 
\usepackage[table,xcdraw]{xcolor}
\usepackage{algorithm}
\usepackage{algpseudocode}
\usepackage{listings} 
\usepackage{inconsolata} 
\usepackage{svg}
\definecolor{codebg}{RGB}{245,245,245}     
\definecolor{codeframe}{RGB}{220,220,220}  
\definecolor{pykw}{RGB}{0,73,168}          
\definecolor{pycm}{RGB}{0,128,128}         
\definecolor{pystr}{RGB}{163,21,21}        
\definecolor{pynum}{RGB}{128,128,128}      

\lstdefinestyle{pyStyle}{
  language=Python,
  backgroundcolor=\color{codebg},
  basicstyle=\linespread{1.0}\ttfamily\footnotesize,
  keywordstyle=\bfseries\color{pykw},
  commentstyle=\itshape\color{pycm},
  stringstyle=\color{pystr},
  numberstyle=\tiny\color{pynum},
  numbers=left,
  numbersep=8pt,
  showstringspaces=false,
  breaklines=true,
  breakatwhitespace=true,
  tabsize=4,
  upquote=true,
  frame=single,
  framerule=0.4pt,
  rulecolor=\color{codeframe},
  columns=fullflexible,
  keepspaces=true,
  captionpos=b
}

\definecolor{cvprblue}{rgb}{0.21,0.49,0.74}
\usepackage[pagebackref,breaklinks,colorlinks,allcolors=cvprblue]{hyperref}

\def\paperID{12313} 
\def\confName{CVPR}
\def\confYear{2026}

\title{Online Data Curation for Object Detection via Marginal Contributions to Dataset-level Average Precision}

\author{Zitang Sun
\and
Masakazu Yoshimura
\and
Junji Otsuka
\and
Atsushi Irie
\and
Takeshi Ohashi \and 
Sony Group Corporation\\
{\tt\small \{Zitang.sun, Masakazu.Yoshimura, Junji.Otsuka, Atsushi.Irie, Takeshi.a.Ohashi\}@sony.com}
}

\begin{document}

\ifMain
    \maketitle
    \vspace{-5mm}
\begin{abstract}
High-quality data has become a primary driver of progress under scale laws, with curated datasets often outperforming much larger unfiltered ones at lower cost. Online data curation extends this idea by dynamically selecting training samples based on the model’s evolving state. While effective in classification and multimodal learning, existing online sampling strategies rarely extend to object detection because of its structural complexity and domain gaps. We introduce DetGain, an online data curation method specifically for object detection that estimates the marginal perturbation of each image to dataset-level Average Precision (AP) based on its prediction quality. By modeling global score distributions, DetGain efficiently estimates the global AP change and computes teacher-student contribution gaps to select informative samples at each iteration. The method is architecture-agnostic and minimally intrusive, enabling straightforward integration into diverse object detection architectures. Experiments on the COCO dataset with multiple representative detectors show consistent improvements in accuracy. DetGain also demonstrates strong robustness under low-quality data and can be effectively combined with knowledge distillation techniques to further enhance performance, highlighting its potential as a general and complementary strategy for data-efficient object detection.
\end{abstract}    
    \vspace{-5mm}
\section{Introduction}
\label{sec:intro}
High-quality data have increasingly become a primary driver of performance in the scale-law era. Across language \cite{gunasekar2023textbooks,liu2024deepseek}, vision \cite{oquab2023dinov2}, and multimodal modeling \cite{abbas2023semdedup,jest,aced}, training on well-curated datasets consistently matches or even exceeds the accuracy of much larger unfiltered datasets, while requiring substantially less compute.
Online data curation takes this a step further by dynamically deciding, at training time, which samples to learn next \cite{loshchilov2015online}. Unlike classic offline dataset pruning or active learning, an online curator adapts to the model’s evolving state, steering optimization toward timely, informative, and non-redundant samples \cite{rholoss}. Recent policies in classification and multimodal contrastive learning implement this idea by ranking samples using a \textit{learnability} signal computed from a teacher and a student, typically the loss gap between a pretrained teacher and the current student \cite{rholoss,evans2024bad}. Intuitively, samples that the teacher handles well (low teacher loss, high prediction quality) but the student struggles with (high student loss, lower prediction quality) represent rich residual knowledge to be learned, and thus are promising training targets.
\begin{figure}[t]
    \centering
    \includegraphics[width=1.0\linewidth]{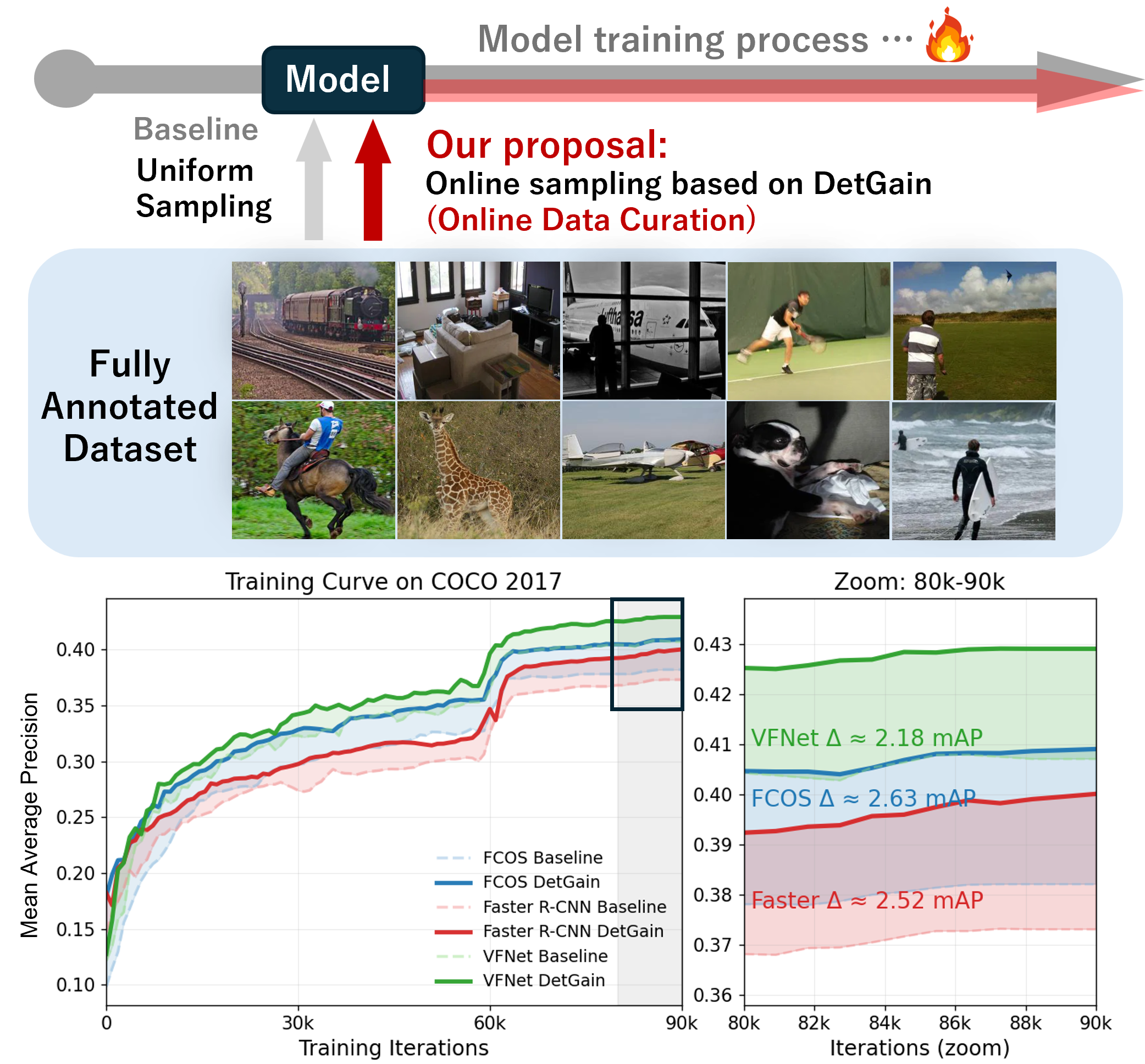}
    \caption{
        Illustration of our online data curation for object detection, which selects
      the most informative samples at each training iteration to boost model training performance.
    }
    \vspace{-4mm}
    \label{fig:csod_difference}
\end{figure}

Despite these advances, online data selection for object detection remains underexplored, and few studies have demonstrated tangible performance gains. This gap persists due to two fundamental challenges. First, defining a consistent score for each image in object detection is inherently complex: a single image may contain no instance or multiple instances, some informative and others noisy or ambiguous \cite{yuan2021multiple,lee2024coreset}. Second, the optimization strategy used in detection destabilizes the loss, which is widely used as a component for online data curation \cite{loshchilov2015online,kawaguchi2020ordered,rholoss,jest,aced}. Detection losses are fragmented across tasks (e.g., classification \cite{fasterrcnn,retinanet}, localization \cite{vfnet}, centerness \cite{fcos}, etc.) and are affected by stochastic proposal sampling and assignment rules. Modules such as RPN sampling \cite{fasterrcnn} or Hungarian matching \cite{detr, deformable,dino} dynamically determine which and how many proposals contribute to the loss. These factors cause loss values to fluctuate and drift across iterations, architectures, and even within the same image; thus, loss-based signals such as “learnability” are unreliable for reflecting the residual knowledge to be learned, making direct adaptation of generic data selection metrics prone to domain gaps.

Our key idea is to score learnability using a metric-aware signal aligned with global (dataset-level) Average Precision (AP) rather than raw loss values. We introduce \textit{\textbf{DetGain}} (Detection Gain), an image-level estimate of each sample’s marginal contribution to the global mAP. For any detector, whether one-stage, two-stage, or transformer-based, the forward pass outputs predicted bounding boxes with confidence scores, labels, and IoUs with ground-truth. By computing DetGain post hoc from these outputs, we avoid reliance on architecture-specific internals and estimate how adding an image perturbs the dataset-level precision–recall curve. However, since mAP is a ranking-based and non-continuous metric, exact per-iteration marginal updates are computationally expensive and noisy. 
To make DetGain practical for online use, we introduce a fast parametric estimator that models true- and false-positive (TP/FP) score distributions (e.g., two-parameter families) and derives an analytic closed form for efficient computation. Learnability is then defined as the {DetGain} gap between the teacher and the student for the same image: when the teacher’s predicted AP contribution exceeds that of the student, the sample is considered informative and prioritized for training, as it still contains residual knowledge to be learned. The method is fully plug-and-play, modifying only the data pipeline without changing model architectures, loss functions, optimizers, or training schedules.


We evaluate DetGain on several representative detectors, including Faster R-CNN \cite{fasterrcnn}, ATSS \cite{atss}, FCOS \cite{fcos}, VFNet \cite{vfnet}, GFL \cite{gfl}, and Deformable DETR \cite{deformable}, on COCO 2017 benchmark \cite{COCO2017}. Acting at a meta-level, DetGain adapts to diverse loss functions and architectures, consistently improving validation performance by up to +2.7 mAP under standard schedules. Moreover, on low-quality datasets with noisy or pseudo labels, DetGain yields gains of up to +6.9 mAP. By sharing the teacher, it can also be combined with knowledge distillation (KD) methods \cite{crosskd,pkd}, further boosting lightweight models.
Existing work has rarely explored image-wise online data sampling that produces measurable gains on standard detection benchmarks. DetGain fills this gap by providing extensive comparisons against alternative online metrics. Our results demonstrate that selecting the most informative samples during training consistently improves both detection accuracy and efficiency (i.e., convergence speed), highlighting the strong potential of DetGain for broader applications in data-efficient learning.

    \section{Related Works}
\label{sec:previousdataset}

\subsection{Online Data Curation}
Online data curation ranks and selects training examples on the fly based on the model’s evolving state~\cite{loshchilov2015online,shah2020choosing,wang2024data,rholoss}. 
Earlier work focuses on hard-example mining, where samples are chosen by high loss~\cite{loshchilov2015online,kawaguchi2020ordered} or prediction uncertainty~\cite{li2006confidence,coleman2019selection}. 
Other variants target examples that are easily forgotten during training~\cite{toneva2018empirical} or reweight losses in a curriculum-learning manner~\cite{jiang2018mentornet,superloss}. 
RAIS~\cite{johnson2018training} introduces importance sampling using normalized gradient magnitudes, while Co-Teaching~\cite{han2018co} trains peer networks to mitigate label noise. 
Xinyue et al.~\cite{hao2025progressive} propose a progressive curriculum that drops data portions over time. 
Recently, RHO-LOSS~\cite{rholoss} formalizes “learnability" via a teacher–student signal emphasizing non-noisy and non-redundant samples; JEST~\cite{jest} extends this to batch-level joint selection for contrastive learning, and ACED~\cite{aced} interprets this batch selection as implicit KD methods.

Despite these advances, few studies have explored data selection metrics tailored to object detection or shown measurable gains through online sampling. 
This gap stems from the complexity of detection models, which involve multiple interdependent objectives and image-level sample definitions with variable instance counts~\cite{liu2020deep}. 
Such properties make generic learnability metrics unreliable because teacher–student loss discrepancies often fail to reflect prediction quality in detection frameworks. 
\textit{DetGain} instead offers a post-hoc, model-agnostic \emph{image} score aligned with dataset-level mAP, ranking images by a teacher-student gap. 
Some existing detectors also highlight the importance of online sampling at the proposal/anchor level~\cite{ohem,ssd,cao2020prime}, e.g., assignment or reweighting schemes such as ATSS and Focal Loss~\cite{atss,retinanet}. However, most of them are architecture-coupled and operate on per-RoI/per-anchor gradients rather than providing an \emph{image-level} selection metric, so their applicability to image-wise online curation is limited. We instead decouple the data flow from model internals, avoiding unstable loss signals and architectural idiosyncrasies, and enabling transfer across detector families. 
Moreover, our method operates within the data pipeline, thus complementing existing internal reweighting or sampling approaches; in our experiments, combining \textit{DetGain} with ATSS or Focal Loss yields consistent additional gains.

\subsection{Active Learning and Offline Coreset Selection}
Early active learning (AL) for object detection mainly relied on uncertainty-based querying, such as entropy or margin scores from detector outputs~\cite{choi2021active,roy2018deep}. 
Later works highlighted the importance of localization quality~\cite{kao2018localization} and combined instance-level cues: CDAL~\cite{agarwal2020contextual} modeled contextual diversity among co-occurring classes, MI-AOD~\cite{yuan2021multiple} estimated instance-wise uncertainty via multiple instance learning, and probabilistic models captured both aleatoric and epistemic uncertainty~\cite{choi2021active}. 
Recent approaches integrate uncertainty and diversity within iterative query loops—Entropy-based AL~\cite{wu2022entropy} combines entropy-guided NMS with prototype diversity, while PPAL~\cite{yang2024plug} ranks samples by calibrated uncertainty and clusters. 
Semi-supervised methods such as Active Teacher~\cite{mi2022active} further couple querying with pseudo-labeling for efficient learning.

In parallel, coreset and dataset selection methods aim to construct compact, representative subsets offline~\cite{sener2017active,borsos2020coresets,mirzasoleiman2020coresets,killamsetty2021grad,lee2024coreset,zhou2024optimizing}.
Most AL and coreset pipelines remain offline or semi-offline and recursive, and the main goal is label efficiency—approaching fully supervised accuracy with minimal human annotations. However, such strategies cannot be directly applied when the dataset is already fully labeled.
In contrast, DetGain performs online, real-time batch selection during training using fully labeled data. 
Instead of handcrafted mixtures of uncertainty or diversity, it uses a teacher–student signal, a dynamic, AP-oriented contribution gap, to identify batches that maximize each image’s marginal contribution to global AP. 
Recent study~\cite{zhou2024optimizing} applies AP-based metrics for offline data selection. DetGain generalizes this concept by dynamically estimating global AP contributions rather than local image-level AP.

    \section{Methods}
\label{sec:methos}

\subsection{Batch-selection Criteria}
\textbf{Previous Learnability definition:}
We denote the model under training as the \textit{student}.  
At each iteration, a ``super-batch'' $\mathcal{S}$ of $B$ samples is first loaded, from which we wish to extract a smaller, more informative \textit{sub-batch}  
$\mathcal{B} = \{x_i\}_{i=1}^{b} \subset \mathcal{S}$  
to perform gradient updates.
To guide this selection, {learnability-based} \cite{rholoss,evans2024bad,jest,aced} scoring functions leverage loss signals from both the \textit{student} model $f_s(x;\theta)$ and a pretrained \textit{teacher} model $f_t(x;\theta^*)$, defined as:
$s_{\text{learn}} = \ell(\mathcal{B}\,|\,f_s) - \ell(\mathcal{B}\,|\,f_t),$
where $\ell(\mathcal{B}\,|\,f)$ denotes the mean loss of model $f$ over batch $\mathcal{B}$.  
Intuitively, samples with high student loss but low teacher loss are prioritized,  
as they indicate regions where the student still lags behind an optimized reference model.
The first term, $\ell(\mathcal{B}\,|\,f_s)$, reflects sample difficulty relative to the learner’s current state.   
Conversely, the second term, $\ell(\mathcal{B}\,|\,f_t)$, favors high-quality examples but ignores the student’s current progress.  
By combining the two, $s_{\text{learn}}$ captures the \textit{improvement margin}, namely how much a sample could help bridge the gap between the student and teacher.
The ratio between sub-batch and super-batch sizes defines the selection ratio $k = b/B$.  
Smaller $k$ values lead to more selective sampling but increase the computational cost of scoring,  
as additional forward passes on the super-batch are required.

\textbf{DetGain-based learnability:}
Loss-based learnability is unstable for object detection because the training loss mixes heterogeneous terms (classification, box regression, etc) with architecture- and scale-dependent weights, and it is not directly aligned with the evaluation metric. We therefore use a \emph{metric-driven} formulation that is detector-agnostic and tied to dataset-level AP.

Let $\mathcal{D}$ be the current dataset (or pool) and $f(\cdot)$ a detector. Let $\mathcal{C}$ be the set of classes and $\mathcal{T}$ the set of IoU thresholds used for evaluation (e.g., COCO uses $\mathcal{T}=\{0.50,0.55,\ldots,0.95\}$). The dataset-level metric is
\begin{equation}
\label{eq:map}
\small
\mathrm{mAP}(f;\,\mathcal{D})
=\frac{1}{|\mathcal{C}|\,|\mathcal{T}|}
\sum_{c\in\mathcal{C}}\sum_{\tau\in\mathcal{T}}
\operatorname{AP}_{c,\tau}\!\big(f;\,\mathcal{D}\big).
\end{equation}

For a candidate image $x\notin\mathcal{D}$, the \emph{DetGain} of $x$ under model $f$ is the marginal perturbation of dataset-level mAP:
\begin{equation}
\small
\label{eq:detgain}
\delta_{\mathrm{mAP}}(x;\,f,\mathcal{D})
\triangleq
\mathrm{mAP}\!\big(f;\,\mathcal{D}\cup\{x\}\big)
-
\mathrm{mAP}\!\big(f;\,\mathcal{D}\big).
\end{equation}
This quantity measures how the true/false positives from $x$, together with their ranks, deform the global precision–recall curve when merged with $\mathcal{D}$.
Let $f_s$ denote the student and $f_t$ a fixed, pretrained teacher. We redefine a learnability score that quantifies the remaining improvement margin on $x$:
\begin{equation}
\label{eq:dg_learn}
s_{\mathrm{DG}}(x)
=
\delta_{\mathrm{mAP}}(x;\,f_t,\mathcal{D})
-
\delta_{\mathrm{mAP}}(x;\,f_s,\mathcal{D}).
\end{equation}
Large values indicate images where the teacher contributes more to global mAP than the student, i.e., where the student still lags in \emph{dataset-level} utility.
Given a super-batch, we score each image with \eqref{eq:dg_learn} and select the top-$b$ to form the sub-batch $\mathcal{B}$. For efficiency, we use the additive approximation
$\delta_{\mathrm{mAP}}(\mathcal{B};\,f,\mathcal{D})
\approx
\sum_{x\in\mathcal{B}}\delta_{\mathrm{mAP}}(x;\,f,\mathcal{D}),$
which is accurate when  $b \ll \mathcal{D}$  and enables per-image, online scoring. Our overall pipeline is shown in Fig. \ref{fig:detgain_pipeline} A.

Directly maximizing $\mathrm{mAP}(f;\mathcal{D})$ during training is notoriously difficult due to the non-smooth and non-decomposable nature of AP and the heavy computational cost across the whole dataset~\cite{aploss1,aploss2}. DetGain preserves the metric target but relocates it to the data selection step, avoiding gradient issues while aligning sampling with evaluation. In the next subsection we introduce a fast estimator for Eq. \eqref{eq:detgain} that supports real-time scoring.

\begin{figure*}[t]
    \centering
    \vspace{-3mm}
    \includegraphics[width=\linewidth]{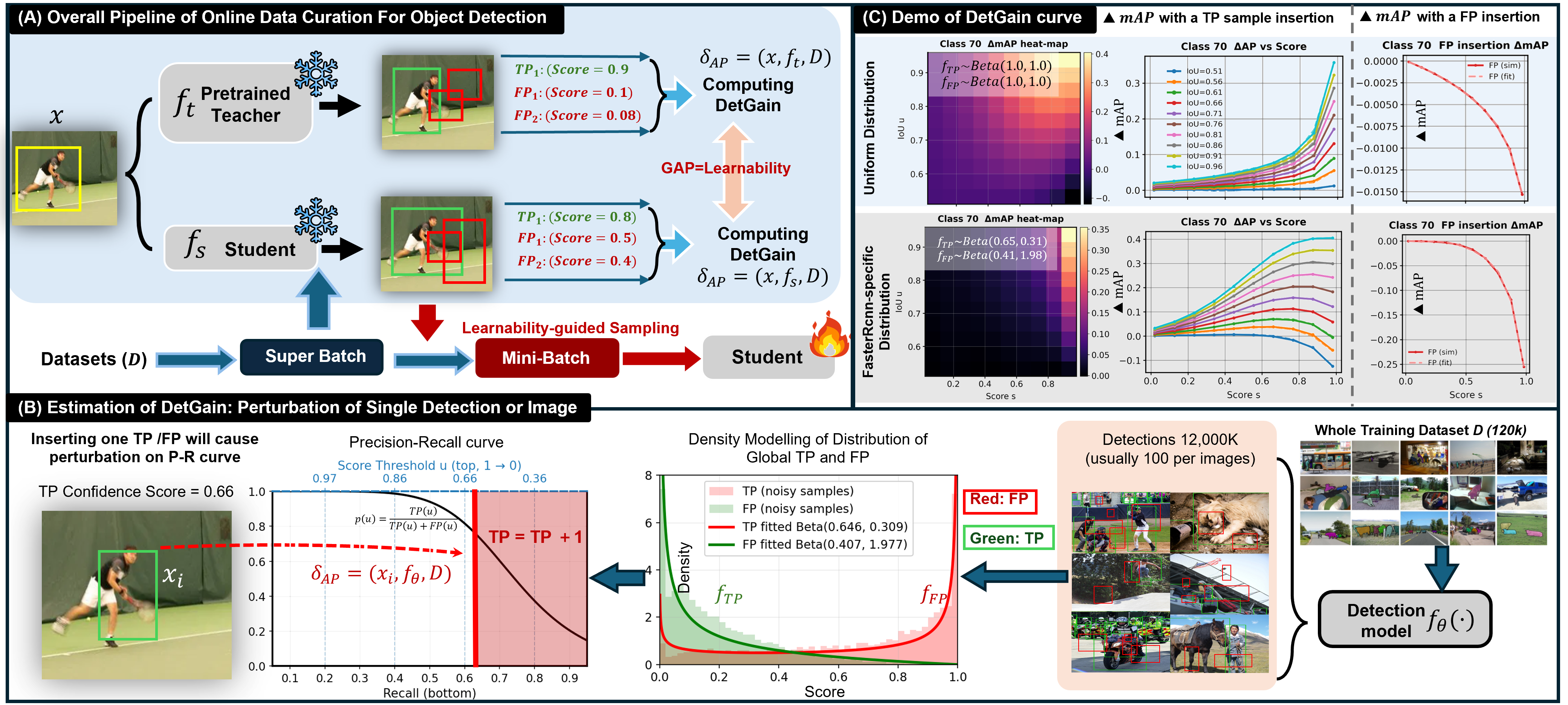}
    \caption{
    \textbf{Overview of the proposed online data curation framework for object detection.} 
    (A) \textit{Overall pipeline:} a pretrained teacher and a student model compute their respective marginal mAP contributions (\(\delta_{\mathrm{AP}}\)) for each image. 
    The difference between them defines the \textit{learnability} score, which guides mini-batch sampling during training. 
    (B) \textit{Estimation of DetGain:} inserting a single TP or FP perturbs the precision–recall curve. 
    We model global TP/FP score distributions and analytically estimate each detection’s contribution to mAP under these densities. 
    (C) \textit{Demonstration of DetGain behavior.}
    {Left:} $\Delta\text{mAP}$ when inserting a class-specific TP detection—determined jointly by the confidence score and bounding-box quality (IoU). {Right:} $\Delta\text{mAP}$ when inserting a class-specific {FP} detection—determined primarily by the prediction score.
    The {bottom row} uses the true TP/FP score distributions measured from Faster R-CNN predictions; recomputing these at every iteration is computationally expensive.
    We adopt the {top row} (uniform distribution) for a model-agnostic faster computation.
    }
    \label{fig:detgain_pipeline}
\end{figure*}

\subsection{Fast Calculation of DetGain}
We aim to approximate the marginal change of dataset-level mAP contributed by inserting the detections from one image without rerunning the full evaluator over the dataset.  
Each detection (predicted bbox) from one image can be viewed as a small perturbation to the global counts of true and false positives $(T,F)$ already accumulated from $\mathcal{D}$. The AP integrates precision along the PR curve as the detection score threshold decreases, making it rank-dependent. To approximate dataset-level $\Delta$\,mAP, we therefore model the score distribution of TPs and FPs in a continuous form.

\textbf{Formulation in the score–threshold domain: }
Let $u\in(0,1)$ be the confidence threshold (scanned from high to low).  
Let $T_{\mathrm{GT}}$ denote the total number of ground truths, and $(T,F)$ the current numbers of true and false positives.  
Let $F_{\mathrm{TP}}(u),F_{\mathrm{FP}}(u)$ be the cumulative distribution functions (CDFs) and $f_{\mathrm{TP}}(u),f_{\mathrm{FP}}(u)$ their probability densities (PDFs).  
The cumulative TP/FP counts above threshold $u$ are:
\begin{equation}
\small
\label{eq:cums}
C_{\mathrm{TP}}(u) = T\,\bigl[1 - F_{\mathrm{TP}}(u)\bigr],\\
C_{\mathrm{FP}}(u) = F\,\bigl[1 - F_{\mathrm{FP}}(u)\bigr],
\end{equation}
and the total predictions above $u$ are $N(u)=C_{\mathrm{TP}}(u)+C_{\mathrm{FP}}(u)$.  
Precision and recall at threshold $u$ are
\begin{equation}
\small
\label{eq:p_r}
p(u) = \frac{C_{\mathrm{TP}}(u)}{N(u)},
r(u) = \frac{C_{\mathrm{TP}}(u)}{T_{\mathrm{GT}}}.
\end{equation}
Hence, the (non-interpolated) AP can be written as the Stieltjes integral:
\begin{equation}
\small
\label{eq:ap-int}
\mathrm{AP}
=\int_{0}^{1} p(u)\,dr(u)
=-\frac{T}{T_{\mathrm{GT}}}\int_{0}^{1} p(u)\,f_{\mathrm{TP}}(u)\,du.
\end{equation}

\textbf{Marginal $\Delta$\,mAP for inserting a detection.}
Consider adding one detection with score $s$. There are two cases.

\emph{(i) when Insert a TP at score $s$.} For $u\le s$, $C_{\mathrm{TP}}'(u)=C_{\mathrm{TP}}(u)+1$ and $N'(u)=N(u)+1$, while $r(u)$ gains a point mass $1/T_{\mathrm{GT}}$ at $u=s$. Using Eq. \eqref{eq:ap-int} one obtains
\begin{equation}
\small
\begin{aligned}
\delta_{\mathrm{AP}}^{\mathrm{TP}}(s)
&= \frac{1}{T_{\mathrm{GT}}}\,\frac{C_{\mathrm{TP}}(s)+1}{N(s)+1}\\
&\quad + \frac{T}{T_{\mathrm{GT}}}\int_{0}^{s}
\frac{C_{\mathrm{FP}}(u)}{N(u)\,[N(u)+1]}\,f_{\mathrm{TP}}(u)\,du.
\label{eq:delta-ap-tp}
\end{aligned}
\end{equation}

The first line is the self-contribution at the insertion point; the second line accounts for precision changes on all existing TPs with scores $\ge s$.

\emph{(ii) when Insert an FP at score $s$.} Recall does not change; for $u\le s$ we have $C_{\mathrm{FP}}'(u)=C_{\mathrm{FP}}(u)+1$ and $N'(u)=N(u)+1$. Thus
\begin{equation}
\small
\delta_{\mathrm{AP}}^{\mathrm{FP}}(s)= -\,\frac{T}{T_{\mathrm{GT}}}\int_{0}^{s}
\frac{C_{\mathrm{TP}}(u)}{N(u)\,[N(u)+1]}\,f_{\mathrm{TP}}(u)\,du.
\label{eq:delta-ap-fp}
\end{equation}
The computational pipeline is illustrated by Fig \ref{fig:detgain_pipeline} (B).

\textbf{Image-level aggregation across classes and IoU thresholds:}
For a single image $x$, under a first-order additive assumption, its marginal mAP change is approximated by summing the per-detection effects after standard one-to-one matching at each IoU threshold $\tau$ and then averaging over classes and thresholds. Generally, COCO-style mAP averages AP over object classes $\mathcal{C}$ and IoU thresholds $\mathcal{T}$ (Eq.~\ref{eq:map}). Accordingly, we account for both the IoU- and class-wise effects in $\delta_{\mathrm{mAP}}(x)$. Let $T_{c,\tau}$ and $F_{c,\tau}$ be the dataset-level totals of TPs and FPs used to compute $\operatorname{AP}_{c,\tau}$, and let $T^{\mathrm{GT}}_c$ be the total number of ground-truth instances of class $c$. For detections on $x$, $\{(s_j,c_j,\mathrm{IoU}_j)\}_{j=1}^{m}$ (score $s_j$, predicted class $c_j$, IoU to the best-matched ground truth), define $y_{j,\tau} \in \{0,1\}$ as the indicator that detection $j$ is a TP at threshold  $\tau$.

Given class/IoU-specific score densities $f^{(c,\tau)}_{\mathrm{TP}}(u)$ and $f^{(c,\tau)}_{\mathrm{FP}}(u)$ (used inside Eqs.~\ref{eq:delta-ap-tp}–\ref{eq:delta-ap-fp}), the image-level marginal is approximated by
\begin{equation}
\small
\label{eq:img-agg}
\begin{aligned}
\delta_{\mathrm{mAP}}(x)
&\approx \frac{1}{|\mathcal{C}|\,|\mathcal{T}|}
  \sum_{c\in\mathcal{C}}\sum_{\tau\in\mathcal{T}}
  \sum_{j:\,c_j=c}
  \Big[
    y_{j,\tau}\,\delta_{\mathrm{AP}}^{\mathrm{TP},(c,\tau)}(s_j)
\\
&\quad
  + \big(1-y_{j,\tau}\big)\,\delta_{\mathrm{AP}}^{\mathrm{FP},(c,\tau)}(s_j)
  \Big],
\end{aligned}
\end{equation}
where $\delta_{\mathrm{AP}}^{\mathrm{TP},(c,\tau)}(\cdot)$ and $\delta_{\mathrm{AP}}^{\mathrm{FP},(c,\tau)}(\cdot)$ are the single-insertion effects from Eqs.~\ref{eq:delta-ap-tp}–\ref{eq:delta-ap-fp}. In practice, we observe that the IoU of matched TP shows an approximately linear relationship with their DetGain. Moreover, since $\delta_{\mathrm{AP}}(\cdot)$ includes the normalization factor $1/T^{\mathrm{GT}}_{c}$ (Eq.~\ref{eq:delta-ap-tp}–\ref{eq:delta-ap-fp}), a TP or FP from a rarer class induces a larger per-class AP change than one from a frequent class. The outer average over classes in mAP further assigns equal weight to each category, thereby naturally mitigating long-tail imbalance.

\textbf{Parametric modeling of score and IoU effects:}
The per-threshold terms in Eq. \eqref{eq:delta-ap-tp}–\eqref{eq:delta-ap-fp} depend on TP/FP score densities for each class and IoU threshold. 
A practical choice is to fit lightweight parametric models from the current student’s predictions, for example using a Beta distribution with method-of-moments fitting, i.e., $u\sim\operatorname{Beta}\!\bigl(\alpha_{\mathrm{TP}},\beta_{\mathrm{TP}}\bigr)$ for TP scores (and analogously for FP), after which the image-level $\Delta\mathrm{mAP}$ is obtained by the one-dimensional numerical integrals in Eq. \eqref{eq:delta-ap-tp}–\eqref{eq:delta-ap-fp}.

To reduce the burden of online distribution updates and keep computation real-time, we adopt a universal simplification where all TP/FP scores follow $\mathrm{Beta}(1,1)$, i.e., a uniform distribution with 
$f^{(c,\tau)}_{\mathrm{TP}}(u)\!\equiv\!f^{(c,\tau)}_{\mathrm{FP}}(u)\!\equiv\!1$ on $(0,1)$. 
This eliminates the need for per-model estimation of the prior-distribution and allows the method to be directly applied to any detector without parameter calibration. More importantly, under the above uniform prior, Eq. \eqref{eq:delta-ap-tp}–\eqref{eq:delta-ap-fp} admit compact closed forms ($s\!\in\!(0,1)$):
\begin{equation}
\small
\begin{aligned}
\delta_{\mathrm{AP}}^{\mathrm{TP}}(s)
&=\frac{1}{T_c^{\mathrm{GT}}}\!
  \left[
  \frac{T(1-s)+1}{A(1-s)+1}
  +\frac{TF}{A^2}\ln\!\frac{A+1}{A(1-s)+1}
  \right],\\[3pt]
\delta_{\mathrm{AP}}^{\mathrm{FP}}(s)
&=-\frac{T^2}{T_c^{\mathrm{GT}}A^2}
  \ln\!\frac{A+1}{A(1-s)+1},
\end{aligned}
\label{eq:fp-unif}
\end{equation}
where \(T=T_{c,\tau}\), \(F=F_{c,\tau}\), and \(A=T+F\).
Both functions are monotonic with respect to $s$—increasing for TPs and decreasing for FPs. 
In deployment, these terms are integrated into the image-level aggregator (Eq.~\ref{eq:img-agg}); 
the analytical solution also enables an $\mathcal{O}(1)$ online cost per detection, 
i.e., {$\mathcal{O}(m)$} per image with $m$ detections.

The uniform prior reflects the maximum-entropy distribution over confidence scores, a random-guess baseline. Under this baseline, DetGain quantifies how much the current detector improves over a naive predictor. We adopt this simplification because (i) it admits a closed-form, $\mathcal{O}(1)$ analytic computation of TP/FP contributions; (ii) it avoids brittle, model-specific density fitting; and (iii) in practice we rank images by the teacher–student DetGain gap, so both models are compared against the same neutral baseline. Although the absolute values are approximate, the induced ordering is stable and informative for sample selection.
We further visualize the numerical solution of Eq.~\eqref{eq:img-agg} as DetGain surfaces over IoU and score (Fig.~\ref{fig:detgain_pipeline} C). We compare two instantiations: (i) the uniform-prior closed forms (Eq.~\eqref{eq:fp-unif}), and (ii) class-/threshold-specific {Beta} fits estimated from Faster R-CNN. Both exhibit the same trend that higher-score TPs produce larger positive gains while negative for FPs—and empirically they deliver nearly identical downstream performance (see Appendix 6 for details).

\begin{figure}[t]
    \centering
    \includegraphics[width=1.0\linewidth]{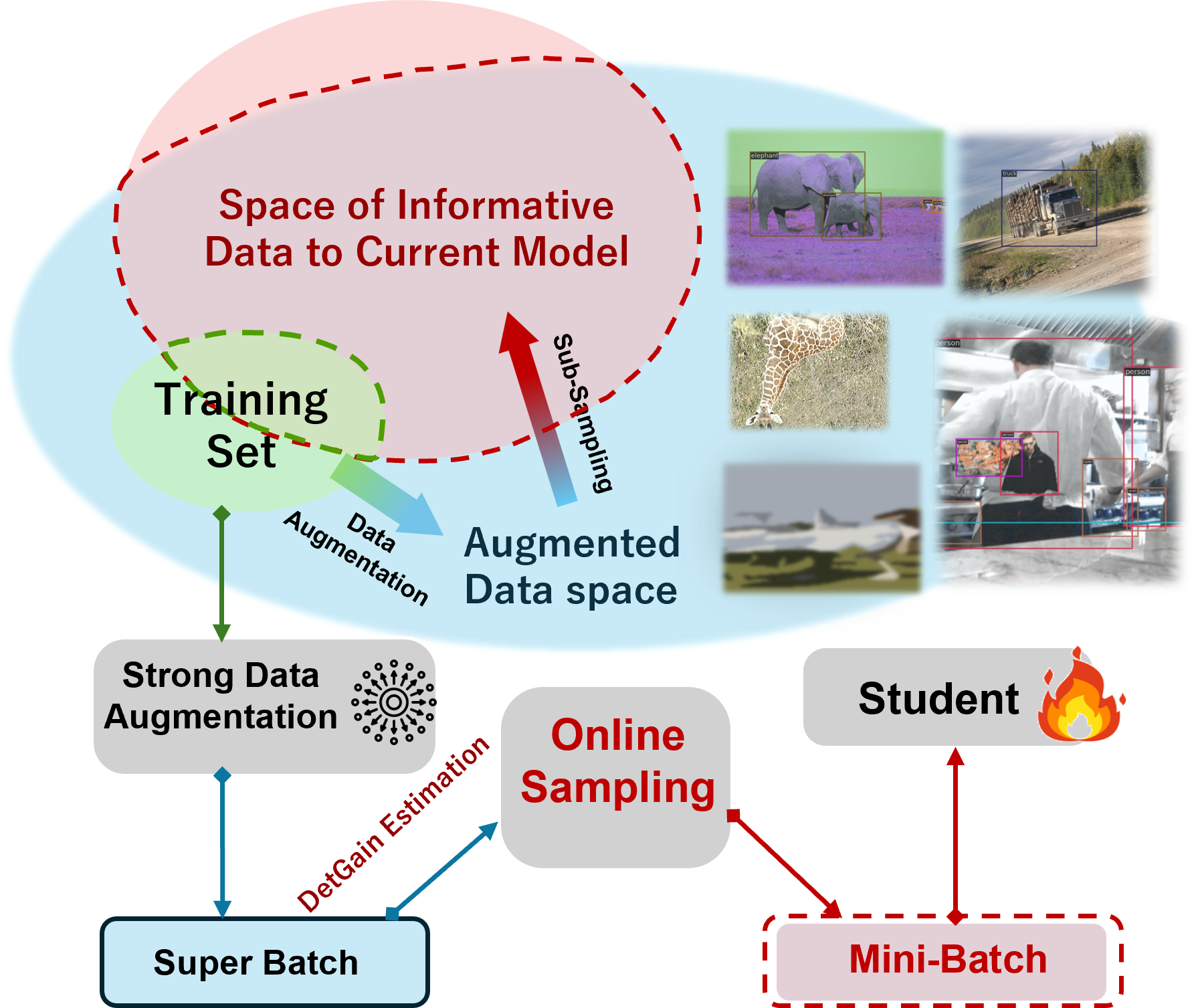}
    \caption{
        \textbf{Joint DetGain–augmentation sampling framework.}
        Strong augmentation expands the \emph{augmented data space} beyond the original training set.
        The subsampling space extends from the small overlap between the training set and informative data 
        ({\color{green!60!black}green}–{\color{red!70!black}red}) to a larger overlap 
        ({\color{blue!70!black}blue}–{\color{red!70!black}red}), reducing overfitting and improving data diversity for the student model.
    }
    \label{fig:aug_sampling_concept}
\end{figure}
\begin{table*}[!ht]
\centering
\footnotesize
\setlength{\tabcolsep}{5pt}
\renewcommand{\arraystretch}{1.25}
\caption{\footnotesize \textbf{Overall results on COCO \texttt{val2017}.}
We report AP@50  and standard AP (AP@[50:95])
“+Data Aug.” applies strong online augmentation.
“+ DetGain” denotes combination of  strong augmentation with DetGain-based online sampling
(subsampling ratio = 20\%).
All student models use ResNet-50 backbones; teachers are same-family models pretrained on the same training set.
(Res152 for CNN-based detectors and Res101 for Deform. DETR). Results are repeated three times with different seeds (mean $\pm$ std).}
\label{tab:main_results}
\begin{tabular}{l l c c c c c c}
\toprule
\textbf{Model} & \textbf{Type} 
& \multicolumn{2}{c}{\textbf{Baseline}}
& \multicolumn{2}{c}{\textbf{+Data Aug.}}
& \multicolumn{2}{c}{\textbf{+DetGain}} \\
\cmidrule(lr){3-4}\cmidrule(lr){5-6}\cmidrule(lr){7-8}
& &  \textbf{AP@50} & \textbf{AP}
& \textbf{AP@50} & \textbf{AP}
& \textbf{AP@50} & \textbf{AP} \\
\midrule
Faster RCNN~\cite{fasterrcnn} & 2-stage, anchor-based 
& 58.3$\pm$.094 & 37.5$\pm$.047
& 58.3$\pm$.125 {\scriptsize\textcolor{gray}{(+0.0)}} & 37.5$\pm$.082 {\scriptsize\textcolor{gray}{(+0.1)}}
& \textbf{61.0}$\pm$\textbf{.125} {\scriptsize\textcolor{gray}{(+2.7)}} & \textbf{40.0}$\pm$\textbf{.125} {\scriptsize\textcolor{gray}{(+2.5)}} \\
\rowcolor{rowgray}
ATSS~\cite{atss} & 1-stage, anchor-based
& 57.3$\pm$.082 & 39.2$\pm$.094
& 56.0$\pm$.141 {\scriptsize\textcolor{gray}{(-1.3)}} & 38.6$\pm$.125 {\scriptsize\textcolor{gray}{(-0.8)}}
& \textbf{59.8}$\pm$\textbf{.094} {\scriptsize\textcolor{gray}{(+2.5)}} & \textbf{41.5}$\pm$\textbf{.047} {\scriptsize\textcolor{gray}{(+2.3)}} \\
FCOS~\cite{fcos} & 1-stage, anchor-free
& 56.7$\pm$.125 & 38.2$\pm$.125
& 56.6$\pm$.094 {\scriptsize\textcolor{gray}{(-0.1)}} & 38.2$\pm$.047 {\scriptsize\textcolor{gray}{(+0.0)}}
& \textbf{59.8}$\pm$\textbf{.125} {\scriptsize\textcolor{gray}{(+3.1)}} & \textbf{41.0}$\pm$\textbf{.047} {\scriptsize\textcolor{gray}{(+2.8)}} \\
\rowcolor{rowgray}
GFL~\cite{gfl} & 1-stage, anchor-based 
& 58.3$\pm$.094 & 40.2$\pm$.082
& 57.9$\pm$.125 {\scriptsize\textcolor{gray}{(-0.4)}} & 40.3$\pm$.094 {\scriptsize\textcolor{gray}{(+0.1)}}
& \textbf{59.9}$\pm$\textbf{.082} {\scriptsize\textcolor{gray}{(+1.6)}} & \textbf{42.0}$\pm$\textbf{.094} {\scriptsize\textcolor{gray}{(+1.8)}} \\
VFNet~\cite{vfnet} & 1-stage, anchor-based 
& 44.3$\pm$.125 & 40.7$\pm$.094
& 44.3$\pm$.082 {\scriptsize\textcolor{gray}{(+0.0)}} & 40.3$\pm$.125 {\scriptsize\textcolor{gray}{(-0.4)}}
& \textbf{46.5}$\pm$\textbf{.094} {\scriptsize\textcolor{gray}{(+2.2)}} & \textbf{42.9}$\pm$\textbf{.082} {\scriptsize\textcolor{gray}{(+2.2)}} \\
\rowcolor{rowgray}
Def. DETR~\cite{deformable} & DETR., anchor-free 
& 65.8$\pm$.141 & 46.6$\pm$.125
& 66.5$\pm$.245 {\scriptsize\textcolor{gray}{(+0.7)}} & 47.5$\pm$.189 {\scriptsize\textcolor{gray}{(+0.9)}}
& \textbf{68.2}$\pm$\textbf{.163} {\scriptsize\textcolor{gray}{(+2.4)}} & \textbf{48.9}$\pm$\textbf{.125} {\scriptsize\textcolor{gray}{(+2.3)}} \\
\bottomrule
\end{tabular}
\end{table*}

\subsection{Combine with Online Data Augmentation}
Pure online sampling can easily overfit. Intuitively, DetGain-based subsampling can be viewed as a pointer oriented toward a hidden subspace containing the most informative data to the current model. However, repeatedly sampling high-learnability samples may cause the training to collapse onto a narrow subspace within the training set.  
Our goal is to improve the global mAP, but such collapse often leads to a gap between training and validation performance (see Table \ref{tab:aug_sampling_ablation}). Previous methods address this by maintaining a hold-out teacher~\cite{rholoss} or expanding the dataset for student~\cite{jest,aced}, both requiring extra data or data partitions.  
We propose a simpler alternative: combining \emph{strong online augmentation} with our image-level DetGain scoring. An augmentation operator 
\(
\mathcal{A}(\cdot) \!\sim\! \lambda
\)
(e.g., color jitter, affine transform, noise, copy-paste) is applied to each sample $x$ before DetGain computation to expand the data space before subsampling.  
This paired design simplifies the RHO-style setup~\cite{rholoss}: the teacher is trained on clean (no augmented) data, while the student learns from augmented views without requiring a hold-out split. Their combination significantly enlarges the sampling space and allows the sampler to filter out low-quality augmentations while focusing on informative regions.  
As shown in Fig.~\ref{fig:aug_sampling_concept}, by first transferring samples into the augmented space and then subsampling toward informative regions, the effective sampling area expands from the small overlap between the training set and informative data (green–red) to a larger overlap with the augmented data space (blue–red).

    \section{Experiments}
\subsection{Implementation Details}
\label{sec:exp-setup}
We evaluate on COCO~2017~\cite{COCO2017} (118k train / 5k val), a widely used and challenging detection benchmark with a long-tailed class distribution, diverse object scales, and dense annotations. To probe robustness, we also study (i) label-noise scenarios, (ii) a pseudo-labeling regime using the \emph{unlabeled} split, and (iii) Extra validation on PASCAL VOC~\cite{voc} and BDD100k~\cite{bdd100k} dataset in appendix.
All experiments are implemented in MMDetection~\cite{mmdetection} (PyTorch) and run on a high-performance computing cluster with NVIDIA A100/H100 GPUs. The effective batch size is 16 for both the baseline and DetGain. For online sampling, each iteration forms an 80-image \emph{super-batch} and backpropagates the top 20\% ranked by the teacher--student DetGain gap (i.e., 16 images).
Unless otherwise noted, we follow MMDetection defaults: a \(1\times\) schedule for CNN-based detectors (\(\approx 90{,}000\) iterations, SGD) and the default 50-epoch schedule for Deformable DETR (\(\approx 3.7{\times}10^{5}\) iterations, AdamW~\cite{adamw}). We also evaluate longer schedules and compare with knowledge-distillation baselines in Sec.~\ref{sec:expkd}. Student models use ResNet-50~\cite{resnet} backbones by default; teachers use the larger backbones (Res101/Res152) pretrained on the same training set by default. All other settings (data preprocessing, optimizer, and learning policy) follow each detector’s default configuration.

\begin{table}[t]
\centering
\footnotesize
\setlength{\tabcolsep}{8pt}
\renewcommand{\arraystretch}{1.1}
\caption{\footnotesize \textbf{Effect of online sampling and data augmentation on Faster R-CNN (COCO val).}
Sampling by learnability alone easily overfits, while strong augmentation alone lacks focus. 
Combining both restores diversity and achieves the best generalization.}
\label{tab:aug_sampling_ablation}
\begin{tabular}{cc|cc}
\toprule
\multicolumn{2}{c|}{\textbf{Setting}} & \multicolumn{2}{c}{\textbf{AP}} \\ 
\cmidrule(lr){1-2}\cmidrule(l){3-4}
\textbf{Data aug.} & \textbf{Online sampling} & \textbf{Train} & \textbf{Val} \\
\midrule
\xmark & \xmark & 44.6 & 37.4 \\ 
\rowcolor{rowgray}
\xmark & \cmark & \textbf{50.3} {\scriptsize\textcolor{gray}{(+5.7)}} & 37.3 {\scriptsize\textcolor{gray}{(-0.1)}} \\
\cmark & \xmark & 40.4 {\scriptsize\textcolor{gray}{(-4.2)}} & 37.5 {\scriptsize\textcolor{gray}{(+0.1)}} \\
\rowcolor{rowgray}
\cmark & \cmark & 45.3 {\scriptsize\textcolor{gray}{(+0.7)}} & \textbf{39.9} {\scriptsize\textcolor{gray}{(+2.5)}} \\
\bottomrule
\end{tabular}
\end{table}

\subsection{Main Results}
As shown in Tab.~\ref{tab:main_results}, DetGain-based online curation consistently improves performance across diverse detector architectures and optimization objectives.
For most models, strong data augmentation alone brings limited mAP gains, as excessive perturbations often degrade sample quality. However, when combined with our sampling pipeline, the model effectively filters out low-quality augmentations, selects more diverse and learnable samples, and achieves substantial improvements.
Overall, across six representative detectors, our method yields an average gain of approximately \(\sim\!+2.0\) mAP. Notably, these improvements are achieved only by modifying the data sampling strategy without any changes to the model architecture, loss function, or training schedule.
\subsection{Ablation Studies}
\label{sec:expablation}
\textbf{Effect of Data Augmentation:}
Sampling purely by high DetGain-based learnability narrows training to a small subspace, causing overfitting: training mAP rises sharply while evaluation mAP barely moves.
Adding strong online augmentation restores diversity and, when combined with online sampling, yields the best generalization (Table~\ref{tab:aug_sampling_ablation}).
The two are complementary: augmentation expands the data space; online curation sub-samples the most informative instances from this enlarged space (Fig.~\ref{fig:aug_sampling_concept}).

\textbf{Teacher Selections.}
We examine the effect of teacher capacity.
Table~\ref{tab:teacher_backbone} shows that larger teacher backbones consistently yield stronger students, echoing \cite{aced} that teacher-driven curation functions as implicit knowledge distillation by providing richer supervision on data quality.
Accordingly, we use a ResNet-152 teacher with the same architecture family as the student by default to maximize capacity-driven benefits.
See Section~\ref{sec:expkd} for further discussion of relationship to KD methods.

\begin{figure}[t!]
    \centering
    \vspace{-3mm}
    \includegraphics[width=0.9\linewidth]{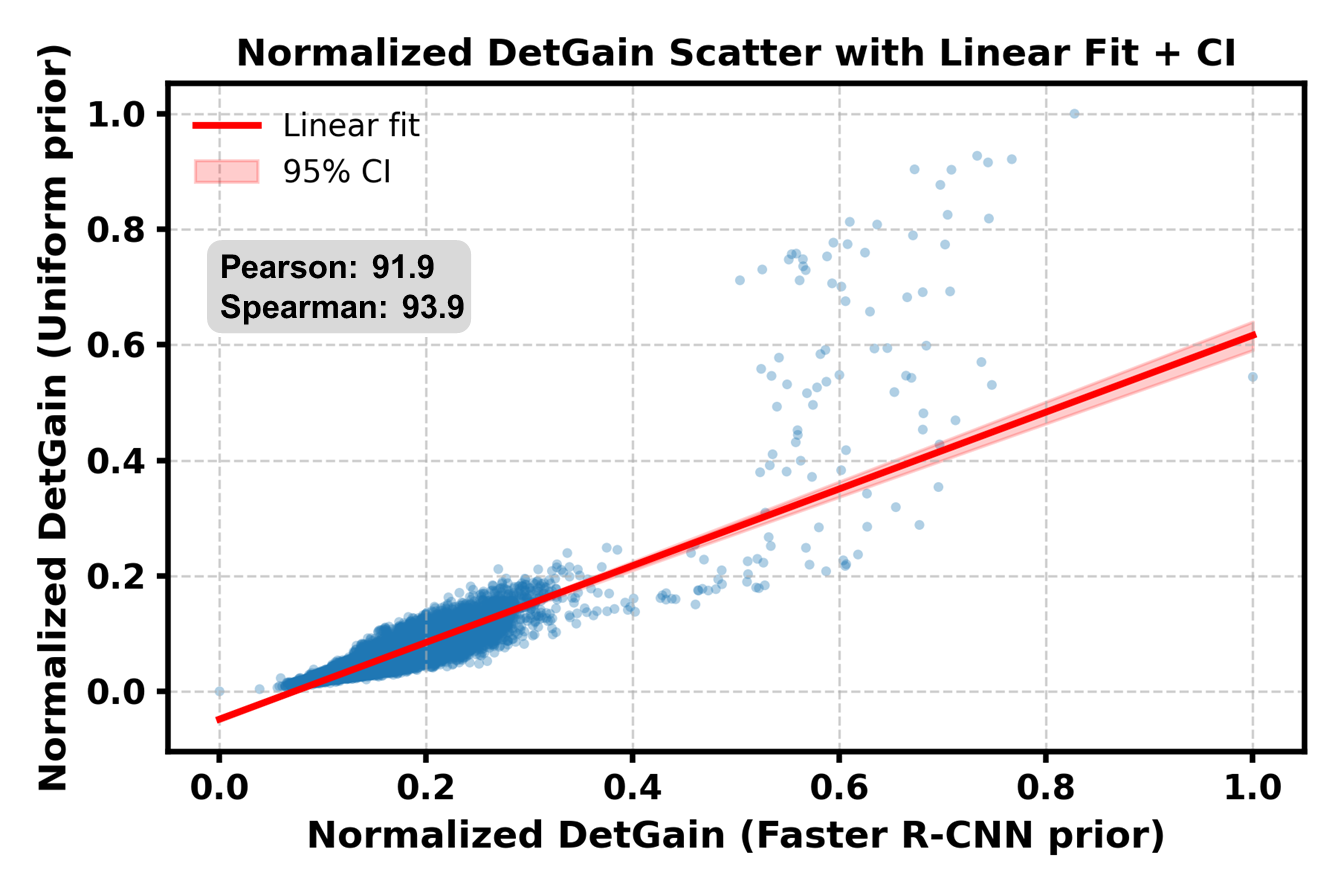}
    \vspace{-2mm}
    \caption{\textbf{Uniform prior v.s. model-fit ranking}. Comparison between DetGain value computed from statistics estimated from Faster R-CNN and uniform prior.}
    \label{detgain_rank}
    \vspace{-2mm}
\end{figure}

\begin{table}[t]
\centering
\footnotesize
\setlength{\tabcolsep}{4.5pt}
\renewcommand{\arraystretch}{1.1}
\caption{\footnotesize \textbf{Teacher backbone used for DetGain scoring.}
All results are mAP on COCO val2017 with ResNet-50 as students' backbone. 
Larger teacher backbones slightly yield stronger student performance.}
\label{tab:teacher_backbone}
\begin{tabular}{l c c c c}
\toprule
\multirow{2}{*}{\textbf{Student}} & \multirow{2}{*}{\textbf{Baseline}} & 
\multicolumn{3}{c}{\textbf{+DetGain (Teacher Backbone)}} \\
\cmidrule(lr){3-5}
& & \textbf{Res50} & \textbf{Res101} & \textbf{Res152} \\
\midrule
Faster R-CNN & 37.4 & 39.6 {\scriptsize\textcolor{gray}{(+2.2)}} & 39.9 {\scriptsize\textcolor{gray}{(+2.5)}} & \textbf{40.0} {\scriptsize\textcolor{gray}{(+2.6)}} \\
\rowcolor{rowgray}
FCOS         & 38.2 & 40.2 {\scriptsize\textcolor{gray}{(+2.0)}} & 40.7 {\scriptsize\textcolor{gray}{(+2.5)}} & \textbf{40.9} {\scriptsize\textcolor{gray}{(+2.7)}} \\
VFNet        & 40.7 & 42.6 {\scriptsize\textcolor{gray}{(+1.9)}} & 42.7 {\scriptsize\textcolor{gray}{(+2.0)}} & \textbf{42.9} {\scriptsize\textcolor{gray}{(+2.2)}} \\
\rowcolor{rowgray}
ATSS & 39.4 & 40.9 {\scriptsize\textcolor{gray}{(+1.5)}} & 41.3 {\scriptsize\textcolor{gray}{(+1.9)}} & \textbf{41.6} {\scriptsize\textcolor{gray}{(+2.2)}} \\
\bottomrule
\end{tabular}
\vspace{-3mm}
\end{table}

\textbf{Effect of Uniform Prior.}
To quantify whether the uniform prior changes the \emph{relative ordering} used for selection, we directly compare
the per-image DetGain rankings produced by (i) the uniform-prior closed form in E.q. \eqref{eq:fp-unif} with fixed TP:FP = 1:9,
and (ii) a detector-specific Beta-fit prior with fitted TP/FP score distributions and a fitted TP:FP ratio.
We simulate a large number of independently sampled super-batches, matching the typical detection setting:
each super-batch contains 64 images, each image produces a fixed budget of 100 post-processed predictions,
and the number of ground-truth instances per image is sampled from the same range as in training;

For each super-batch, we compute the Spearman rank correlation $\rho$ between the 64 uniform-prior DetGain values
and the 64 Beta+ratio-fit DetGain values, and then aggregate $\rho$ over all trials.
Across hundreds of trials, the resulting ranking agreement is consistently high (mean Spearman $\rho \approx 0.94$; Fig.~\ref{detgain_rank}),
supporting that the uniform prior provides a stable monotonic surrogate for sample selection.


\begin{figure}[t]
    \centering
    \vspace{-3mm}
    \includegraphics[width=1.0\linewidth]{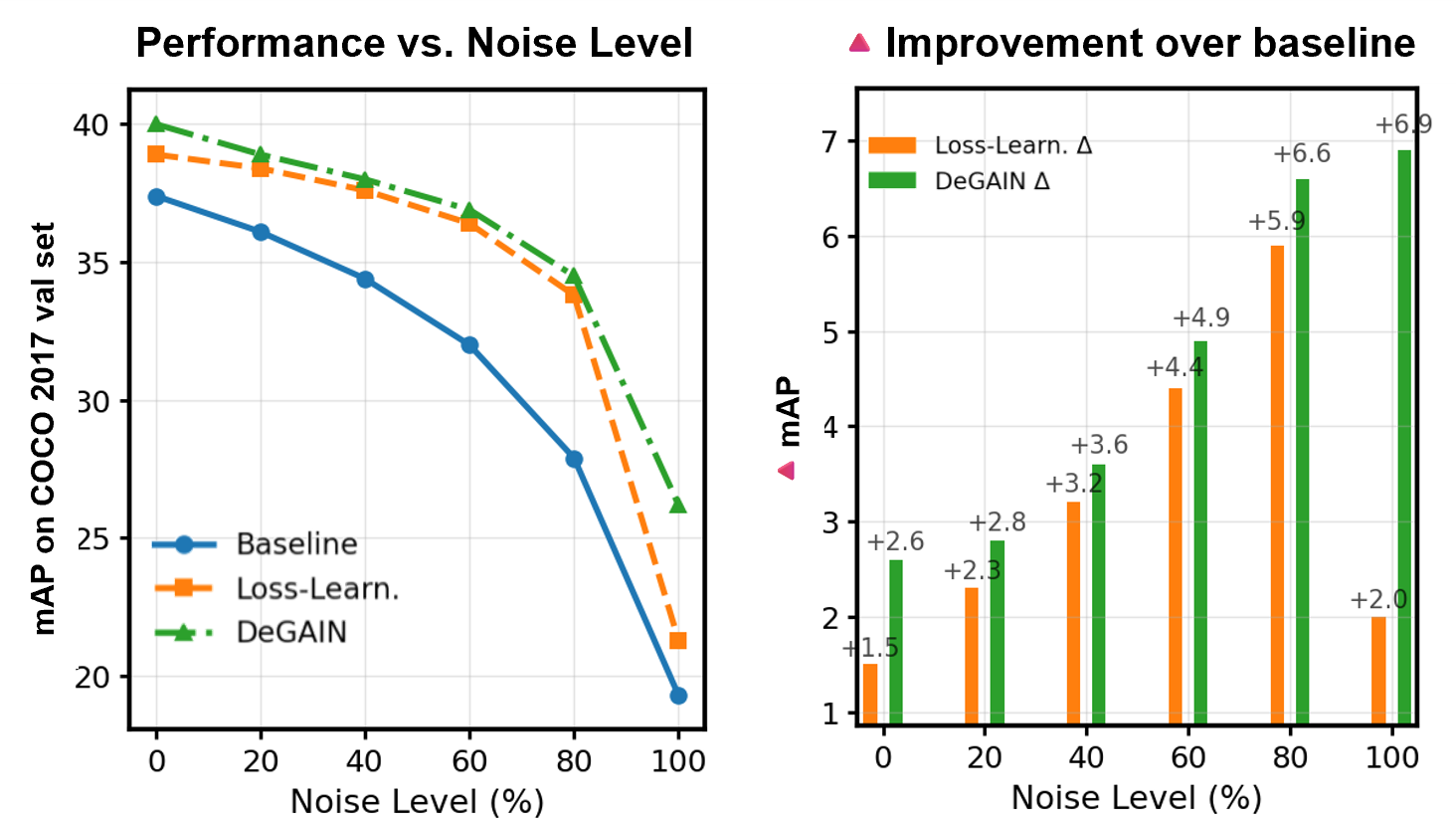}
    \caption{\textbf{Robustness under noisy annotations.}
    Results with \textit{Faster R-CNN–Res50} (stu.) and \textit{Faster R-CNN–Res152} (tea.) on COCO~2017 under controlled annotation-noise ratios.
    DetGain remains more stable than loss-based learnability baselines~\cite{rholoss,jest,evans2024bad} and the uniform baseline across varying noise levels.}
    \vspace{-3mm}
    \label{fig:noisycurve}
\end{figure}

\begin{table*}[ht]
\centering
\footnotesize
\caption{\footnotesize \textbf{Comparison of online sampling metrics on COCO val 2017.}
We report COCO Val AP for three representative students. Each method selects the top 20\% samples from a 64-image super-batch with data augmentation.
DetGain shows consistent and stable improvements across all architectures, whereas loss- or gradient-based metrics exhibit large variations due to their dependency on optimization design.
For a fair comparison, all selection metrics adopt the same online-selection style from the super-batch of each iteration.}
\label{tab:compare_metrics}
\setlength{\tabcolsep}{7pt}
\renewcommand{\arraystretch}{1.25}
\begin{tabular}{lcccccccccccc}
\toprule
& \multicolumn{4}{c}{\textbf{Faster R-CNN\,--\,Res50}} 
& \multicolumn{4}{c}{\textbf{FCOS\,--\,Res50}} 
& \multicolumn{4}{c}{\textbf{ATSS\,--\,Res50}} \\
\cmidrule(lr){2-5}\cmidrule(lr){6-9}\cmidrule(lr){10-13}
\textbf{Sampling metric} 
& $\mathbf{AP_s}$ & $\mathbf{AP_m}$ & $\mathbf{AP_l}$ & \textbf{AP}
& $\mathbf{AP_s}$ & $\mathbf{AP_m}$ & $\mathbf{AP_l}$ & \textbf{AP}
& $\mathbf{AP_s}$ & $\mathbf{AP_m}$ & $\mathbf{AP_l}$ & \textbf{AP} \\
\midrule
Uniform (baseline)                 & 21.3 & 40.9 & 48.5 & 37.3 & 22.9 & 41.8 & 48.2 & 38.2 & 23.3 & 43.2 & 49.6 & 39.4\\
\rowcolor{rowgray}
GradNorm~\cite{johnson2018training}          & 22.1 & 41.5 & 47.7 & 37.4 & 23.7 & 42.0 & 47.7 & 38.4 & 24.2 & 43.1 & 49.4 & 39.3 \\
Loss (hard mining)~ \cite{loshchilov2015online}    & 23.7 & 40.3 & 44.7 & 36.3 & 20.9 & 37.9 & 43.8 & 34.5 & 24.7 & 41.6 & 44.6 & 37.7 \\
\rowcolor{rowgray}
Cls. Entropy~\cite{mi2022active}~    & 19.6 & 38.4 & 47.5 & 34.8 & 23.1 & 40.6 & 43.2 & 36.6 & 23.5 & 40.6 & 47.7 & 37.0\\
Entropy, Info., Diversity~\cite{mi2022active}                 & 22.6 & 41.4 & 47.8 & 37.5 & 22.4 & 40.8 & 45.9 & 37.1 & 22.7 & 41.7 & 48.3 & 38.2\\
\rowcolor{rowgray}
Image-AP~\cite{zhou2024optimizing}           & 23.3 & 42.3 & 49.8 & 38.3 & 24.2 & 43.4 & 50.7 & 39.4 & 24.7 & 43.6 & 51.5 & 40.0\\
Loss-learnability~\cite{rholoss,jest} & 23.6 & 42.5 & 50.1 & 38.9 & 22.0 & 41.4 & 49.6 & 38.1 & 25.9 & 44.3 & 51.5 & 40.4\\
\rowcolor{rowgray}
\textbf{DetGain (ours)}             & \textbf{24.4} & \textbf{44.1} & \textbf{50.8} & \textbf{40.0} 
& \textbf{26.0} & \textbf{45.0} & \textbf{51.1} & \textbf{40.9}
& \textbf{26.1} & \textbf{45.5} & \textbf{53.1} & \textbf{41.6}\\
\bottomrule
\end{tabular}
\end{table*}

\textbf{Robustness under Noisy Annotations.}
We inject annotation noise in the training set by (1) adding fake boxes, (2) removing valid ones, (3) perturbing box locations, and (4) label swaps, controlling the overall noise ratio as the fraction of affected GT boxes.
Fig.~\ref{fig:noisycurve} summarizes the results across noise levels: DetGain outperforms both the uniform baseline and loss-based learnability~\cite{rholoss,jest,evans2024bad}.
This robustness stems from scoring each image by its contribution to global mAP, which promotes cleaner, informative samples while down-weighting corrupted ones.
In contrast, loss-based criteria, which directly depend on unstable per-sample losses, are less effective at identifying high-quality data, especially when noise level approaches 100\%.
See Appendix 2 for extra analysis on pseudo-label scenarios.

\subsection{Comparison with Other Methods} Table~\ref{tab:compare_metrics} compares our DetGain-based data curation with various existing sampling metrics on COCO val2017.
In addition to loss- and gradient-based indicators commonly used in early studies~\cite{loshchilov2015online,johnson2018training}, we include recent loss-learnability approaches~\cite{rholoss,jest,aced}, as well as entropy-based and hybrid entropy–diversity criteria widely applied 

\begin{table}[h]
\setlength{\tabcolsep}{10pt}
\renewcommand{\arraystretch}{1.25}
\centering
\caption{\footnotesize \textbf{Complementarity between DetGain and Knowledge Distillation.}
Results on FCOS–Res50~\cite{fcos}. DetGain achieves performance comparable to modern KD methods~\cite{pkd,crosskd} and can be combined with them for further improvement. Unlike KD, DetGain requires no architectural or loss modifications and is less sensitive to teacher strength.}
\vspace{-1mm}
\label{tab:kd_combo}
\footnotesize
\begin{tabular}{ l c c c}
\toprule
\textbf{Teacher} & \textbf{DetGain} & \textbf{KD} & \textbf{AP} \\
\midrule
N.A             & $\times$ & $\times$   & 38.5 \\
\midrule
\rowcolor{rowgray}
FCOS-Res50    & $\checkmark$ & $\times$   & 40.8 {\scriptsize\textcolor{gray}{(+2.3)}} \\
\rowcolor{rowgray}
FCOS-Res101   & $\checkmark$ & $\times$   & 41.1 {\scriptsize\textcolor{gray}{(+2.6)}} \\

\midrule
FCOS-Res50    & $\times$ & PKD        & 38.2 {\scriptsize\textcolor{gray}{(-0.3)}} \\
FCOS-Res101   & $\times$ & PKD        & 40.9 {\scriptsize\textcolor{gray}{(+2.4)}} \\

\midrule
\rowcolor{rowgray}
 FCOS-Res50    & $\times$ & CrossKD    & 39.2 {\scriptsize\textcolor{gray}{(+0.7)}} \\
 \rowcolor{rowgray}
 FCOS-Res101   & $\times$ & CrossKD    & 41.1 {\scriptsize\textcolor{gray}{(+2.6)}} \\

\midrule
 FCOS-Res101   & $\checkmark$ & PKD        & 41.7 {\scriptsize\textcolor{gray}{(+3.2)}} \\
 \textbf{FCOS-Res101} & $\checkmark$ & \textbf{CrossKD} & \textbf{42.2} {\scriptsize\textcolor{gray}{(+3.7)}} \\
\bottomrule
\end{tabular}
\vspace{-2mm}
\end{table}
in active learning for object detection~\cite{wu2022entropy,mi2022active}. We also incorporate the image-wise AP metric proposed by~\cite{zhou2024optimizing}.
All methods are run in the same online framework with batch-wise sampling and strong augmentation (not offline selection). 
For image-wise AP, we define a learnability signal as the teacher–student difference in per-image AP. 
Unlike this discrete, local metric, DetGain estimates each image’s continuous marginal contribution to \emph{global} mAP.

Empirically, DetGain delivers higher absolute accuracy and markedly \emph{more stable} gains across diverse detectors and optimization objectives. 
Loss-/gradient-based strategies (e.g., hard mining, GradNorm) vary with the detector’s internal loss scaling and dynamics, and can fluctuate when switching across anchor-based vs. anchor-free or one- vs. two-stage designs. 
By directly targeting dataset-level mAP, DetGain is architecture-agnostic and remains consistent across loss formulations and model families.
\vspace{-1mm}
\subsection{Discussion with Knowledge Distillation}
\label{sec:expkd}
Tab.~\ref{tab:kd_combo} further explores the relationship between DetGain-based curation and existing KD methods that utilize the pretrained teacher~\cite{pkd,crosskd, lkd, mulkd}.
With FCOS–Res50 under a $2\times$ schedule (matching the setting of CrossKD), DetGain alone is comparable with state-of-the-art KD methods (e.g., PKD~\cite{pkd}, CrossKD~\cite{crosskd}). 
Unlike KD methods, which modify the loss function or introduces additional training heads and feature-alignment modules, our method is \emph{non-intrusive}: it operates purely on the data level without altering the model architecture, training objective, or inference procedure. This simplicity allows DetGain to be easily applied to any detector without code-level modification.

Compared with KD methods, DetGain is less sensitive to the strength of the teacher. Even when the teacher and student share the same architecture (e.g., ResNet-50), DetGain remains effective, whereas KD performance drops significantly with weaker teachers.
Importantly, DetGain and KD are \emph{complementary}: KD transfers feature-level knowledge, while DetGain enhances sample-level quality. Combining the two is equal to training the student on the most informative samples while aligning the internal representations.

    \vspace{-1mm}
\section{Conclusion and Limitations}
\vspace{-1mm}
For object detection, we propose the first data-efficient online curation pipeline that pairs strong augmentation with DetGain-based subsampling to select informative samples aligned with the model’s current state. 
DetGain is model-agnostic and consistently improves multiple detector families, while remaining robust under noisy annotations, pseudo-labels, and when combined with KD methods.
Limitations include added training time from the pre-sampling stage. Moreover, the data augmentation used in this study is relatively naïve and could be further improved. We leave these directions for future investigation toward more efficient and adaptive online training for object detection task.

\label{sec:Conclusion}

    {
        \small
        \bibliographystyle{ieeenat_fullname}
        \bibliography{main}
    }
\fi

\ifSup
    \clearpage
\setcounter{page}{1}
\maketitlesupplementary
\newcommand{\LNA}{\ln\!\frac{A+1}{A(1-s)+1}}

\paragraph{\textbf{ Appendix 1. Ablation studies on teacher with asynchronized architecture.}}
\label{sec:S1}
We tested FasterRcnn-Res50 with teachers with asynchronous backbones or model families.
We found that the performance drops if  the teacher does not remain within the same model family. As shown in Table~\ref{tab:async_teachers}, asynchronous curation across very different architectures often yields weaker gains. This observation echoes the knowledge-distillation literature, where architectural mismatch leads to a knowledge gap between teacher and student. Hence, we generally select the same architecture but a larger backbone (Res101/Res152) as the default teacher for data curation.
\begin{table}[htbp]
\centering
\caption{\textbf{Asynchronous teacher-student pairing.}
Different teacher architectures yield weaker gains than same-family teachers (student: Faster R-CNN–Res50, baseline 37.4).}
\label{tab:async_teachers}
\small
\begin{tabular}{l c c}
\toprule
\textbf{Teacher (for scoring)} & \textbf{Teacher mAP} & \textbf{Student mAP} \\
\midrule
Faster RCNN–ResX101 \cite{resnetxt}& 42.5 & 38.9 \\
\rowcolor{rowgray}
Faster RCNN–SwinL \cite{swin} & 49.4 & 38.6 \\
Faster RCNN–Res101 \cite{fasterrcnn} &  41.8 & \textbf{39.9} \\
\rowcolor{rowgray}
Deform. DETR–Res50 \cite{deformable}   & 46.2 & 38.1 \\
DINO–Res50 \cite{dino}         &  49.0& 38.3 \\
\bottomrule
\end{tabular}
\end{table}

\paragraph{\textbf{ Appendix 2. Ablation studies on low quality data.}}
Section 4.3 in the main text analyzes DetGain under artificially corrupted training data.
To evaluate robustness under low-quality data, we construct noisy variants of the COCO training set following a simple corruption pipeline. For each image, noise is injected with probability \(p \in \{0.2,\,0.4,\,0.6,\,0.8,\,1.0\}\). We apply four corruption types:
\begin{itemize}
  \item {Bounding box jittering}: randomly scale each box \([x,y,w,h]\) by factors \(s_w, s_h \in [1\!-\!0.5,\,1\!+\!0.5]\), enforcing a minimum deviation of \(5\%\). The box is re-centered and clipped to the image.
  \item {Label noise}: uniformly replace the category of a random subset of boxes, affecting a fraction drawn from \([0.2,\,0.5]\) of instances in the image.
  \item {Deletion}: randomly remove a fraction \([0.2,\,0.5]\) of ground-truth boxes.
  \item {Fake box addition}: insert \(m\) synthetic boxes with \(m \sim \lfloor n \cdot r \rfloor\), \(r \in [0.2,\,0.5]\), size sampled as \(w \in [0.05W,\,0.2W]\), \(h \in [0.05H,\,0.2H]\), and reject if \(\max_{a} \mathrm{IoU}(\mathrm{box}, a) \ge 0.1\). We cap at 20 fake boxes per image.
\end{itemize}
Each \(p\) yields an independent annotation file generated in an offline manner.

We further evaluate the DetGain on pseudo-labeled datasets by incorporating 120k additional COCO-unlabeled images \cite{COCO2017}.
Specifically, we use YOLOv8-nano \cite{yolov8} to generate pseudo-labels for the unlabeled images and adopt a Faster R-CNN-Res152 teacher to perform data curation while training a Faster R-CNN-Res50 student.
As summarized in Table \ref{tab:pseudo_label}, DetGain again yields consistent improvements under these automatically generated, noisy labels.
When the pseudo-labeled data are combined with the clean annotations, performance further increases—resembling a semi-supervised learning regime.
However, we simply include the pseudo-labeled samples in the curated training set without employing any specialized semi-supervised learning techniques.
This result indicates that DetGain can flexibly cope with both random annotation noise and model-induced label noise.
Notably, we intentionally avoid using the same architecture for pseudo-label generation and for student training, since identical models tend to share systematic errors that are difficult to disentangle during data scoring.

\begin{table}[htbp]
\centering
\small
\caption{\textbf{Pseudo-labeling results with YOLOv8 \cite{yolov8}.} 
Teacher for data curation: Faster R-CNN–Res152. 
Student: Faster R-CNN–Res50. 
DetGain improves both clean and pseudo-labeled data, showing robustness to model-generated label noise.}
\label{tab:pseudo_label}
\setlength{\tabcolsep}{7pt}
\begin{tabular}{l c c c}
\toprule
\textbf{Dataset} & \textbf{Baseline} & \textbf{+Aug} & \textbf{+DetGain} \\
\midrule
Train & 37.4 & 37.5 & \textbf{40.0} {\scriptsize\textcolor{gray}{(+2.6)}} \\
\rowcolor{rowgray}
Unlabeled & 32.8 & 32.6 & \textbf{35.6} {\scriptsize\textcolor{gray}{(+2.8)}} \\
Unlabeled + Train & 37.6 & 36.6 & \textbf{40.6} {\scriptsize\textcolor{gray}{(+3.0)}} \\
\bottomrule
\end{tabular}
\end{table}

\paragraph{\textbf{Appendix 3. Additional mathematical derivations}}
\label{app:math}

\begin{table}[t]
\centering
\footnotesize
\setlength{\tabcolsep}{6pt}
\renewcommand{\arraystretch}{1.25}
\begin{tabular}{p{0.25\linewidth} p{0.65\linewidth}}
\toprule
\textbf{Symbol} & \textbf{Definition / Domain} \\
\midrule
$u\in(0,1)$ & Score threshold scanned from high to low. \\
\rowcolor{rowgray}
$\mathcal{D}$ & Current dataset / pool used to accumulate counts. \\
$T_{\mathrm{GT}}\in\mathbb{N}$ & Total number of ground-truth instances over $\mathcal{D}$. \\
\rowcolor{rowgray}
$T,F\in\mathbb{N}$ & Dataset-level counts of true/false positives already accumulated from $\mathcal{D}$. \\
$A=T+F$ & Total number of positives (TP+FP) accumulated so far. \\
\rowcolor{rowgray}
$F_{\mathrm{TP}}(u)$, $f_{\mathrm{TP}}(u)$ & CDF/PDF of TP scores (w.r.t.\ $u$). \\
$F_{\mathrm{FP}}(u)$, $f_{\mathrm{FP}}(u)$ & CDF/PDF of FP scores (w.r.t.\ $u$). \\
\rowcolor{rowgray}
$C_{\mathrm{TP}}(u)$ & Cumulative TPs \emph{above} threshold $u$:
$\;C_{\mathrm{TP}}(u)=T\,[1-F_{\mathrm{TP}}(u)]$. \\
$C_{\mathrm{FP}}(u)$ & Cumulative FPs \emph{above} threshold $u$:
$\;C_{\mathrm{FP}}(u)=F\,[1-F_{\mathrm{FP}}(u)]$. \\
\rowcolor{rowgray}
$N(u)$ & Total predictions above $u$:
$\;N(u)=C_{\mathrm{TP}}(u)+C_{\mathrm{FP}}(u)$. \\
$p(u)$ & Precision at $u$: $\;p(u)=\dfrac{C_{\mathrm{TP}}(u)}{N(u)}$. \\
\rowcolor{rowgray}
$r(u)$ & Recall at $u$: $\;r(u)=\dfrac{C_{\mathrm{TP}}(u)}{T_{\mathrm{GT}}}$. \\
$\mathcal{C},\mathcal{T}$ & Class set and IoU-threshold set for mAP aggregation. \\
\bottomrule
\end{tabular}
\caption{Notation used in Appendix~A3 (dataset-level scoring and AP integration).}
\label{tab:a3-notation}
\end{table}

\noindent\textbf{A3.1 Preliminaries and notation.}
For simplicity, we define all notations in Table \ref{tab:a3-notation}.
The (non-interpolated) AP has the Stieltjes form
\begin{equation}
\label{eq:app-ap-int-stieltjes}
\mathrm{AP}=\int_0^1 p(u)\,dr(u).
\end{equation}
Since $r(u)=\tfrac{T}{T_{\mathrm{GT}}}\bigl[1-F_{\mathrm{TP}}(u)\bigr]$,
we have $dr(u)=-(T/T_{\mathrm{GT}})\,f_{\mathrm{TP}}(u)\,du$, yielding
\begin{equation}
\label{eq:app-ap-int}
\mathrm{AP}=-\frac{T}{T_{\mathrm{GT}}}\int_0^1 p(u)\,f_{\mathrm{TP}}(u)\,du.
\end{equation}

\medskip
\noindent\textbf{A3.2 Single-insertion $\Delta$AP: general derivation.}
Insert a \emph{single detection} with score $s$ into the ranked list. Denote
the updated precision/recall by $p'(u),r'(u)$ and $\mathrm{AP}'=\int_0^1 p'(u)\,dr'(u)$.
Decompose
\begin{equation}
\begin{aligned}
\label{eq:app-dap-split}
\delta\mathrm{AP}
=\mathrm{AP}'-\mathrm{AP}
=\underbrace{\int_0^1 p'(u)\,d~\!\big[r'(u)-r(u)\big]}_{\text{(I)}}\\
+\underbrace{\int_0^1 \big[p'(u)-p(u)\big]\,dr(u)}_{\text{(II)}}.
\end{aligned}
\end{equation}
Term (I) accounts for the \emph{impulse} in recall if a TP is inserted at $u=s$;
term (II) accounts for \emph{precision reweighting} on the existing TP measure $dr(u)$.

\paragraph{Case (i): insert a TP with score $s$.}
For $u\le s$, $C'_{\mathrm{TP}}=C_{\mathrm{TP}}+1$ and $N'=N+1$; for $u>s$ nothing changes.
Recall gains a point mass $1/T_{\mathrm{GT}}$ at $u=s$:
\begin{equation}
\label{eq:app-dr-jump}
d\big[r'(u)-r(u)\big]=\frac{1}{T_{\mathrm{GT}}}\,\delta(u-s)\,du.
\end{equation}
Hence the \emph{self-contribution} is
\begin{equation}
\label{eq:app-self-tp}
\text{(I)}=\frac{1}{T_{\mathrm{GT}}}\,p'(s)
=\frac{1}{T_{\mathrm{GT}}}\,\frac{C_{\mathrm{TP}}(s)+1}{N(s)+1}.
\end{equation}
For term (II), only $u\le s$ matters. Using
\begin{equation}
\small
\label{eq:app-dp-tp}
\begin{aligned}
p'(u)-p(u)
&=\frac{C_{\mathrm{TP}}(u)+1}{N(u)+1}-\frac{C_{\mathrm{TP}}(u)}{N(u)}\\
&=\frac{N(u)-C_{\mathrm{TP}}(u)}{N(u)\,[N(u)+1]}
 =\frac{C_{\mathrm{FP}}(u)}{N(u)\,[N(u)+1]},
\end{aligned}
\end{equation}
and $dr(u)=-(T/T_{\mathrm{GT}})f_{\mathrm{TP}}(u)\,du$, we obtain
\begin{equation}
\label{eq:app-prec-shift-tp}
\text{(II)}
=\frac{T}{T_{\mathrm{GT}}}
\int_{0}^{s}\!
\frac{C_{\mathrm{FP}}(u)}{N(u)\,[N(u)+1]}\,f_{\mathrm{TP}}(u)\,du.
\end{equation}
Combining \eqref{eq:app-self-tp}–\eqref{eq:app-prec-shift-tp}:
\begin{equation}
\begin{aligned}
\label{eq:app-delta-ap-tp-final}
\delta_{\mathrm{AP}}^{\mathrm{TP}}(s)
&=\frac{1}{T_{\mathrm{GT}}}\,\frac{C_{\mathrm{TP}}(s)+1}{N(s)+1}
+ \\
&\frac{T}{T_{\mathrm{GT}}}\!
\int_{0}^{s}\!
\frac{C_{\mathrm{FP}}(u)}{N(u)\,[N(u)+1]}\,f_{\mathrm{TP}}(u)\,du.
\end{aligned}
\end{equation}

\paragraph{Case (ii): insert an FP with score $s$.}
Recall does not change, so (I)$=0$. For $u\le s$, $C'_{\mathrm{TP}}=C_{\mathrm{TP}}$ and $N'=N+1$,
\begin{equation}
\small
\label{eq:app-dp-fp}
p'(u)-p(u)
=\frac{C_{\mathrm{TP}}(u)}{N(u)+1}-\frac{C_{\mathrm{TP}}(u)}{N(u)}
=-\frac{C_{\mathrm{TP}}(u)}{N(u)\,[N(u)+1]}.
\end{equation}
Thus
\begin{equation}
\label{eq:app-delta-ap-fp-final}
\delta_{\mathrm{AP}}^{\mathrm{FP}}(s)
=-\frac{T}{T_{\mathrm{GT}}}
\int_{0}^{s}\!
\frac{C_{\mathrm{TP}}(u)}{N(u)\,[N(u)+1]}\,f_{\mathrm{TP}}(u)\,du.
\end{equation}

\medskip
\noindent\textbf{A3.3 Closed forms under a uniform prior (Beta(1,1)).}
For general applicability across detectors and efficient scoring, we adopt the following uniform-prior simplification:
\begin{equation}
\label{eq:app-uniform-prior}
f^{(c,\tau)}_{\mathrm{TP}}(u)\equiv f^{(c,\tau)}_{\mathrm{FP}}(u)\equiv 1,\qquad
T_{c,\tau}=T_c^{\mathrm{GT}},\qquad
\end{equation}
For a fixed $(c,\tau)$, write $T=T_{c,\tau}$, $F=F_{c,\tau}$, $A=T+F$.
Then:
\begin{equation}
\label{eq:app-cums-uniform}
\left\{
\begin{aligned}
C_{\mathrm{TP}}(u)&=T(1-u),\\
C_{\mathrm{FP}}(u)&=F(1-u),\\
N(u)&=A(1-u).
\end{aligned}
\right.
\end{equation}
\paragraph{(1) TP insertion: closed form for \eqref{eq:app-delta-ap-tp-final}.}
Split the two terms:
\[
\small
\delta_{\mathrm{AP}}^{\mathrm{TP}}(s)
=\underbrace{\frac{1}{T_{\mathrm{GT}}}\frac{C_{\mathrm{TP}}(s)+1}{N(s)+1}}_{\text{self term}}
\;+\;
\underbrace{\frac{T}{T_{\mathrm{GT}}}\int_0^s
\frac{C_{\mathrm{FP}}(u)}{N(u)\,[N(u)+1]}\,du}_{\text{precision reweighting}}.
\]

\emph{Self term.} Substitute $C_{\mathrm{TP}}(s)=T(1-s)$ and $N(s)=A(1-s)$:
\begin{equation}
\small
\label{eq:tp-self-uniform-correct}
\frac{1}{T_{\mathrm{GT}}}\,\frac{T(1-s)+1}{A(1-s)+1}.
\end{equation}

\emph{Precision reweighting term.} Algebraic simplification gives
\[
\small
\frac{C_{\mathrm{FP}}(u)}{N(u)\,[N(u)+1]}
=\frac{F}{A}\cdot\frac{1}{A(1-u)+1}.
\]
Hence
\[
\small
\int_0^s \frac{C_{\mathrm{FP}}(u)}{N(u)\,[N(u)+1]}\,du
=\frac{F}{A}\int_0^s \frac{du}{A(1-u)+1}.
\]
Let $t=1-u$ ($dt=-du$). As $u:0\to s$, $t:1\to 1-s$:
\[
\small
\int_0^s \frac{du}{A(1-u)+1}
=\int_{1-s}^{1}\frac{dt}{At+1}
=\frac{1}{A}\,\LNA.
\]
Therefore
\begin{equation}
\small
\label{eq:tp-int-uniform-correct}
\frac{T}{T_{\mathrm{GT}}}\int_0^s \frac{C_{\mathrm{FP}}(u)}{N(u)\,[N(u)+1]}\,du
=\frac{TF}{T_{\mathrm{GT}}A^2}\,\LNA.
\end{equation}

Combine \eqref{eq:tp-self-uniform-correct}–\eqref{eq:tp-int-uniform-correct}:
\begin{equation}
\small
\label{eq:tp-closed-uniform}
\boxed{\;
\begin{aligned}
\delta_{\mathrm{AP}}^{\mathrm{TP}}(s)
=\frac{1}{T_{\mathrm{GT}}}\!\left[
\frac{T(1-s)+1}{A(1-s)+1}
+\frac{TF}{A^2}\,\LNA
\right]
\;
\end{aligned}
}
\end{equation}

\paragraph{(2) FP insertion: closed form for \eqref{eq:app-delta-ap-fp-final}.}
Similarly,
\[
\frac{C_{\mathrm{TP}}(u)}{N(u)\,[N(u)+1]}
=\frac{T}{A}\cdot\frac{1}{A(1-u)+1}.
\]
Thus
\[
\small
\int_{0}^{s}\frac{C_{\mathrm{TP}}(u)}{N(u)\,[N(u)+1]}\,du
=\frac{T}{A^2}\,\LNA.
\]
Plug into \eqref{eq:app-delta-ap-fp-final}:
\begin{equation}
\small
\label{eq:fp-closed-uniform}
\boxed{\;
\delta_{\mathrm{AP}}^{\mathrm{FP}}(s)
=-\frac{T^2}{T_{\mathrm{GT}}A^2}\,\LNA\
\;}
\end{equation}

\noindent

Both \eqref{eq:tp-closed-uniform}--\eqref{eq:fp-closed-uniform} are $O(1)$ to evaluate and remain
\textit{monotonic} in the detection score $s$---increasing for true positives (TPs) and decreasing for false positives (FPs).
We adopt a \textit{uniform prior} over $(0,1)$ as a maximum-entropy, model-agnostic baseline, which yields closed-form weights
without per-iteration density fitting.
Importantly, our method is used for \emph{ranking}: samples are selected by the teacher--student DetGain gap $s_{\mathrm{DG}}(x)$,
so mild bias in absolute DetGain values largely cancels when taking the difference, while the desired qualitative dependencies are preserved
(higher-scoring TPs contribute more positively; higher-scoring FPs contribute more negatively; and rarer classes with smaller $T_c^{\mathrm{GT}}$
produce proportionally larger per-class shifts).

Further simplification is possible by fixing a global TP:FP ratio (e.g., $T{:}F=1{:}9$ on COCO) rather than re-estimating $(T,F)$ per detector.
This is motivated by the common practice that modern detectors output a fixed number of predictions per image (e.g., 100),
while COCO images contain roughly $\sim$10\% ground-truth instances per prediction budget on average.
Crucially, any fixed $T{:}F$ ratio only rescales the same monotonic functions in \eqref{eq:tp-closed-uniform}--\eqref{eq:fp-closed-uniform}
and therefore does not affect the \emph{relative ordering} that drives selection.

\begin{figure}[t!]
    \centering
    \includegraphics[width=1.0\linewidth]{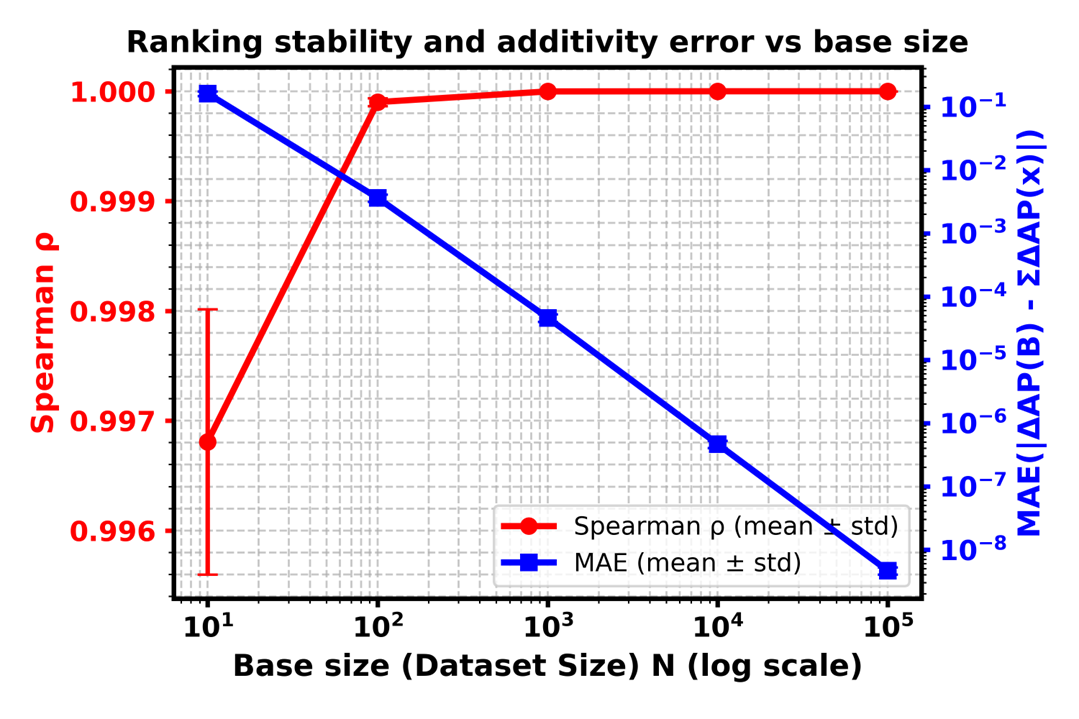}
    \vspace{-2mm}
   \caption{\textbf{Ranking stability and additivity error vs.\ base set size.}
We simulate the super-batch selection setting (e.g., $64\!\rightarrow\!16$) using fitted detector outputs at a real checkpoint.
For each candidate batch $B$, we compare the additive estimate $\hat{s}(B)=\sum_{x\in B}\Delta AP(x)$ with the true batch gain
$s(B)=AP(\mathcal{D}\cup B)-AP(\mathcal{D})$.
As the base dataset size $|\mathcal{D}|$ increases, ranking agreement quickly approaches 1 (left axis: Spearman $\rho$, mean$\pm$std),
while the additivity error decreases (right axis: $\mathrm{MAE}(|\Delta AP(B)-\sum_x \Delta AP(x)|)$, mean$\pm$std).}
    \label{fig:detgain_rank}
    \vspace{-2mm}
\end{figure}

\noindent\textbf{A3.4 Assumptions and first-order additivity.}
Our batch score uses a first-order expansion of the evaluation metric around the current base set $\mathcal{D}$:
for a small selected subset $\mathcal{B}$ from a large super-batch,
\begin{equation}
\label{eq:first_order_additivity}
\delta_{\mathrm{mAP}}(\mathcal{B}; f,\mathcal{D})
=\sum_{x\in\mathcal{B}}\delta_{\mathrm{mAP}}(x; f,\mathcal{D})
+O\!\left(\frac{|\mathcal{B}|}{|\mathcal{D}|}\right).
\end{equation}
In our setting, $\mathcal{B}$ is formed by ranking candidates inside a small sub-batch
(e.g., selecting $k{=}20\%$ from a super-batch), while $\mathcal{D}$ is the full training set (typically orders of magnitude larger),
thus $|\mathcal{B}|\!\ll\!|\mathcal{D}|$ and the remainder term is negligible \emph{for ranking}.

To quantify the magnitude of ignored interactions, we simulate the exact super-batch selection protocol used in training
(e.g., $64\!\rightarrow\!16$) using fitted detector outputs at a real checkpoint.
For each candidate batch $B$, we compare the additive estimate
$\hat{s}(B)=\sum_{x\in B}\Delta AP(x)$ with the true batch gain
$s(B)=AP(\mathcal{D}\cup B)-AP(\mathcal{D})$.
As $|\mathcal{D}|$ increases, the interaction becomes rapidly negligible:
the ranking agreement between $\hat{s}(B)$ and $s(B)$ quickly approaches 1
(Spearman $\rho \approx 1$; see Fig.~\ref{fig:detgain_rank}).
This supports the validity of using Eq.~\eqref{eq:first_order_additivity} as a \emph{ranking surrogate} within each super-batch.
DetGain is computed from \emph{post-processed} predictions (after each detector's decoding and NMS),
so intra-image non-linear interactions induced by NMS are already reflected in the final TP/FP set and their confidence scores.
The sanity check in Fig.~\ref{fig:detgain_rank} also indicates that any residual cross-sample interaction
does not destabilize the batch ranking in the regime where $|\mathcal{B}|\!\ll\!|\mathcal{D}|$.

\textbf{Appendix 4. Verification via Monte Carlo simulation}
\begin{figure}[t]
\centering
\includegraphics[width=\linewidth]{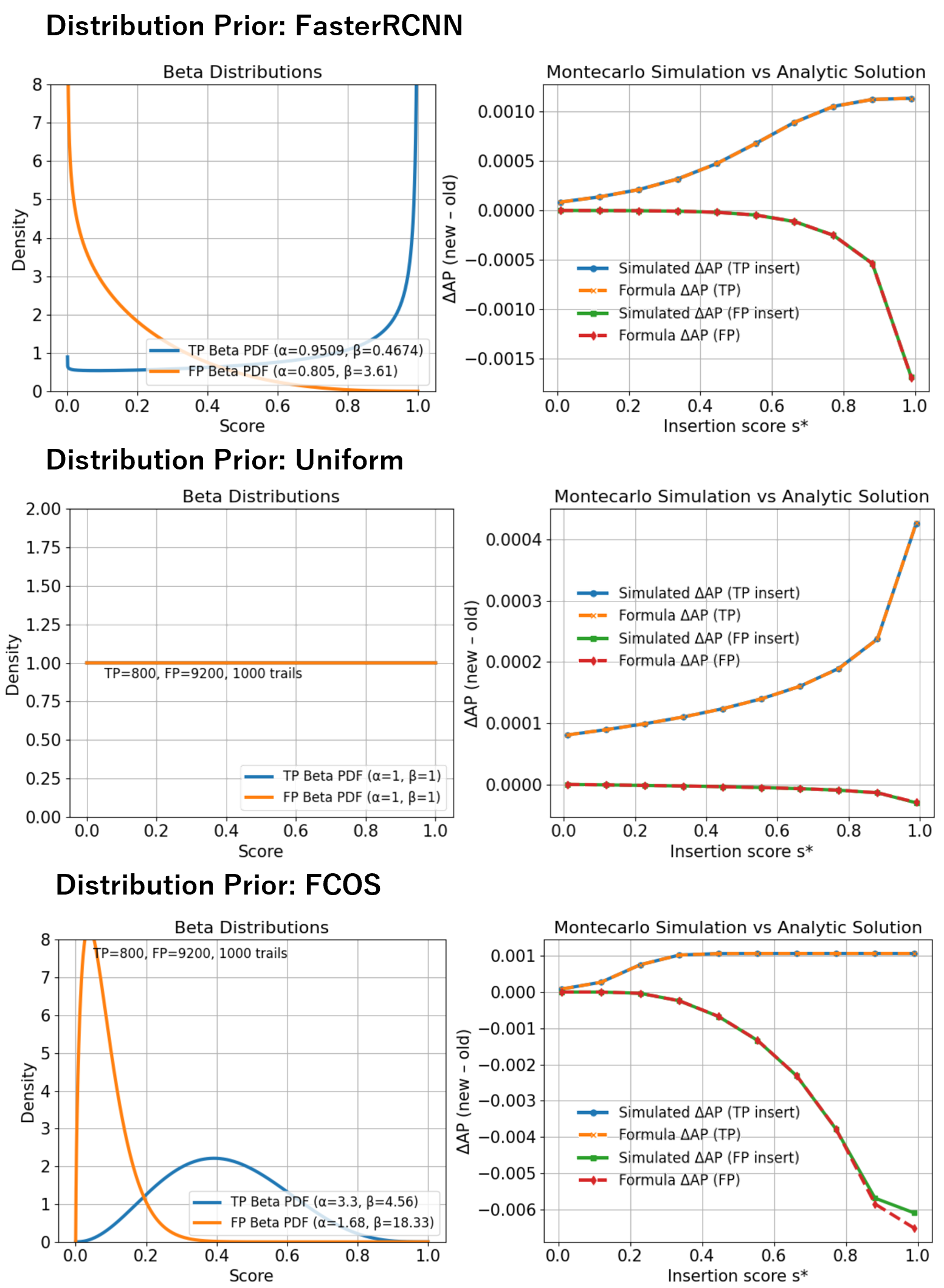}
\caption{\textbf{Monte Carlo verification of the analytic $\Delta$AP formulation.}
Each subplot compares simulated and analytic $\Delta$AP for TP and FP insertions under different Beta priors (top: Faster R-CNN, bottom: FCOS).
The excellent agreement demonstrates the correctness and numerical stability of our closed-form derivation.}
\label{fig:motedetgain}
\end{figure}

To verify the correctness of our analytic formulation in Eq.~(\ref{eq:app-delta-ap-tp-final}, \ref{eq:app-delta-ap-fp-final}), we perform a Monte Carlo simulation that directly emulates the insertion of a single detection under controlled TP/FP score distributions.
For each setting of detector priors (e.g., Faster R-CNN or FCOS), we draw TP and FP scores from their respective Beta distributions, repeatedly compute dataset-level AP before and after inserting one detection with a fixed score~$s^*$, and record the resulting $\Delta$AP.
By averaging over thousands of random trials, the stochastic estimate converges to the expected ground-truth value at the statistical level.
Specifically, we first estimate the Beta distributions of TP and FP scores from each detector trained on the COCO training set and use these distributions as priors for simulation. In each run, we draw \(T=800\) TP scores and \(F=9200\) FP scores
(\(A=T+F\)) to mimic a similar scenario to COCO.
The total number of ground-truth instances is fixed to
\(T_{\mathrm{GT}}=1000\).
We evaluate insertion scores \(s^*\in[0.01,0.99]\) at 10 evenly spaced values.
For each \(s^*\), we repeat the Monte Carlo procedure for 1000 independent
trials and record the mean \(\mathbb{E}[\Delta\mathrm{AP}]\) for both cases:
(i) \emph{TP insertion} (adding a correct detection of score \(s^*\)),
and (ii) \emph{FP insertion} (adding a spurious detection of score \(s^*\)).

Fig. \ref{fig:motedetgain} compares the Monte Carlo results with our analytic solution obtained from Eq.~(\ref{eq:app-delta-ap-tp-final}–\ref{eq:app-delta-ap-fp-final}).
The two curves almost perfectly overlap across all score ranges and detectors, confirming that the proposed closed forms faithfully capture the expected marginal change in AP.
While Monte Carlo simulation requires extensive sampling—thousands of random draws per score and hundreds of repetitions—our analytic expression only needs a lightweight one-dimensional numerical integration.
In practice, using roughly 300 uniformly sampled points is sufficient to obtain a numerically stable $\Delta$AP estimate within $10^{-4}$ of the Monte Carlo result, offering over three orders of magnitude faster computation.

\paragraph{\textbf{Appendix 5. Implementation details.}}
We implement \textit{DetGain} as a thin, model-agnostic wrapper around the training loop. The curator does not modify the detector per se. It only (i) queries a standard \texttt{predict(...)} interface on a \emph{super-batch}, (ii) computes per-image metric-driven \emph{DetGain} for both student and teacher, and (iii) slices/reorders the super-batch into a \emph{sub-batch} for the actual gradient step. Because DetGain consumes only common detection outputs (boxes, scores, labels, and matched IoUs), it works across one-stage/two-stage/transformer detectors, across different codebases (e.g., MMDetection/Detectron2), and with heterogeneous teacher-student pairs.

\textit{Teacher choice.}
We require a pretrained teacher optimized on the same training set. In principle, the teacher can come from any architecture family, independent of the student. In practice, we find that using a teacher from the same family with a stronger backbone (e.g., ResNet-101/152), and trained with same schedules gives the most stable gains.

\textit{Data pipeline and augmentation.}
The data loader applies the baseline pipeline (resize/flip/crop/photometric, etc.) to produce a \emph{strong-augmented} super-batch of size \(B\).
DetGain selection is performed on this super-batch to obtain indices of the top-\(k=\lfloor B\cdot\rho\rfloor\) images, \(\rho\in(0,1]\).
The per-iteration procedure is summarized in Algorithm~\ref{alg:detgain-iter}.


\medskip

\begin{algorithm}[t]
\caption{Online DetGain Curation (model-agnostic, per iteration)}
\label{alg:detgain-iter}
\begin{algorithmic}[1]
\Require ratio \(\rho\in(0,1]\); student \(f_s\); pretrained/optional teacher \(f_t\);
DetGain estimator \(\mathcal{G}\); super-batch size \(B\).
\State \textbf{Data pipeline (strong aug):}\; load super-batch \(\{(x_i,\mathrm{GT}_i)\}_{i=1}^{B}\).
\State \textbf{No-grad \& AMP:}\; stop gradient recording and enable mixed precision.
\State \textbf{Student prediction:}\; set \(f_s\) to \texttt{eval}; \(\mathrm{pred}^s \gets f_s.\mathrm{predict}(\{x_i\})\). \Comment{No GT passed to \texttt{predict}}
\State \textbf{Student DetGain:}\; \(g^s \gets \mathcal{G}(\mathrm{pred}^s,\{\mathrm{GT}_i\}) \in \mathbb{R}^{B}\).
\State \textbf{Teacher prediction:}\; set \(f_t\) to \texttt{eval}; \(\mathrm{pred}^t \gets f_t.\mathrm{predict}(\{x_i\})\).
  \State \textbf{Teacher DetGain:}\; \(g^t \gets \mathcal{G}(\mathrm{pred}^t,\{\mathrm{GT}_i\}) \in \mathbb{R}^{B}\).
\State \textbf{DetGain-based learnability:}\; \(\ell_i \gets g^t_i - g^s_i,\; i=1,\dots,B\).
\State \textbf{Select top-\(k\):}\; \(k=\max(1,\lfloor \rho B \rfloor)\), \(\mathcal{I} \gets \mathrm{TopK}(\ell, k)\).
\State \textbf{Re-enable grad \& exit AMP.}
\State \textbf{Form sub-batch:}\; \(\tilde{x}=\{x_i\}_{i\in\mathcal{I}},\;\tilde{\mathrm{GT}}=\{\mathrm{GT}_i\}_{i\in\mathcal{I}}\). 
\State \textbf{Train step:}\; set \(f_s\) to \texttt{train}; \(\mathcal{L} \gets f_s.\mathrm{loss}(\tilde{x},\tilde{\mathrm{GT}})\); backprop \& optimize.
\end{algorithmic}
\end{algorithm}




\begin{table*}[!h]
\centering
\caption{\textbf{Summary of strong data augmentation components.}
Each operator is independently sampled with the listed probability (\%).
Probabilities are adjusted by detector family: one-stage (lower), two-stage (default), and transformer-based (higher).
To enlarge the augmentation space, we combine three branches: plain (37.5\%), strong (37.5\%), and Copy--Paste (25\%).}
\label{tab:aug_details}
\renewcommand{\arraystretch}{1.1}
\resizebox{\linewidth}{!}{
\begin{tabular}{l l c c c}
\toprule
\textbf{Category} & \textbf{Augmentation} & \textbf{Branch} & \textbf{Typical Param.} & \textbf{Prob. (\%)} \\
\midrule
\multirow{5}{*}{\textbf{Geometric}}
  & RandomCrop (MinIoU)   & Copy--Paste, Strong & Crop by IoU threshold range $[0.1, 0.9]$ & 20--30 \\
  & RandomResize          & All                 & Multi-scale resize, short side $480$--$960$ px & 100 \\
  & RandomFlip            & All                 & Horizontal or vertical flip ($p=0.5$) & 50 \\
  & RandomAffine          & Strong              & Rotation $\pm 10^{\circ}$, scale $[0.8, 1.2]$, shear $\pm 6^{\circ}$ & 20 \\
  & Copy--Paste           & Copy--Paste, Strong & Paste up to 7 instances per image with valid masks & 25--40 \\
\midrule
\multirow{6}{*}{\textbf{Photometric}}
  & Brightness / Contrast & All                 & Intensity and contrast shift ($\pm 0.3$) & 30--40 \\
  & Hue--Saturation--Value& All                 & Color tone shift ($\pm 0.3$) & 20 \\
  & CLAHE                 & Strong              & Local contrast equalization (clip limit $=2.0$) & 15 \\
  & Posterize             & Strong              & Reduce bit depth (2--5 bits/channel) & 10--15 \\
  & Color Distortion      & Strong              & Standard MMDet photometric distortion & 30--50 \\
  & CoarseDropout         & Strong              & Random black patches ($\le 10\%$ area) & 20--25 \\
\midrule
\multirow{5}{*}{\textbf{Noise / Blur / Compression}}
  & Gaussian Blur         & Strong / Copy--Paste & Blur kernel size 5--15 & 10--20 \\
  & Motion Blur           & Strong / Copy--Paste & Simulated camera motion (5--15 px) & 10--20 \\
  & GaussNoise            & Strong               & Additive Gaussian noise (variance 10--40) & 20--30 \\
  & ISO / Multiplicative Noise & Strong         & Sensor/exposure noise ($\alpha\in[0.8, 1.2]$) & 15--20 \\
  & Image Compression     & Strong / Copy--Paste & JPEG quality $[40, 95]$ & 15--20 \\
\bottomrule
\end{tabular}
}
\end{table*}

To improve data diversity and generalization, we adopt a unified augmentation framework that integrates geometric, photometric, and compositional transformations.  
All augmentations are implemented within the MMDetection pipeline, combining \texttt{Albumentations}~\cite{albu} and \texttt{Copy-Paste} modules.  
The same augmentation pool is shared across all detectors, while the application probabilities and strengths are roughly adjusted per model family, as we observed different detectors exhibit different generalization abilities under data augmentation.  
In particular, one-stage detectors are generally less robust to strong transformations, so we apply lighter augmentations; two-stage detectors use moderate settings; and transformer-based detectors, which demonstrate higher augmentation tolerance, are trained with stronger configurations.  
Table~\ref{tab:aug_details} summarizes the main augmentation operators and their typical strength and application probabilities.

For one-stage detectors (e.g., FCOS, ATSS, SSD), geometric perturbations are disabled or downscaled to preserve localization stability.  
Two-stage detectors (e.g., Faster R-CNN) adopt the default probabilities listed in Table~\ref{tab:aug_details}.  
Transformer-based detectors (e.g., Deformable DETR) use the full-strength configuration, introducing stronger geometric transformations and increasing the Copy–Paste and color distortion probabilities by up to 1.2×.  
This adaptive tuning ensures each detector family receives an appropriately strong yet stable augmentation regime while maintaining a consistent data distribution across experiments.  
Note that these augmentations remain relatively naïve and are not extensively optimized per model, given the vast hyperparameter space and the dynamic data selection driven by DetGain-based subsampling, which automatically emphasizes informative samples.  
More adaptive approaches such as online auto-augmentation or conditional data generation could serve as promising directions for future exploration.

\begin{figure*}[h]
\centering
\includegraphics[width=\linewidth]{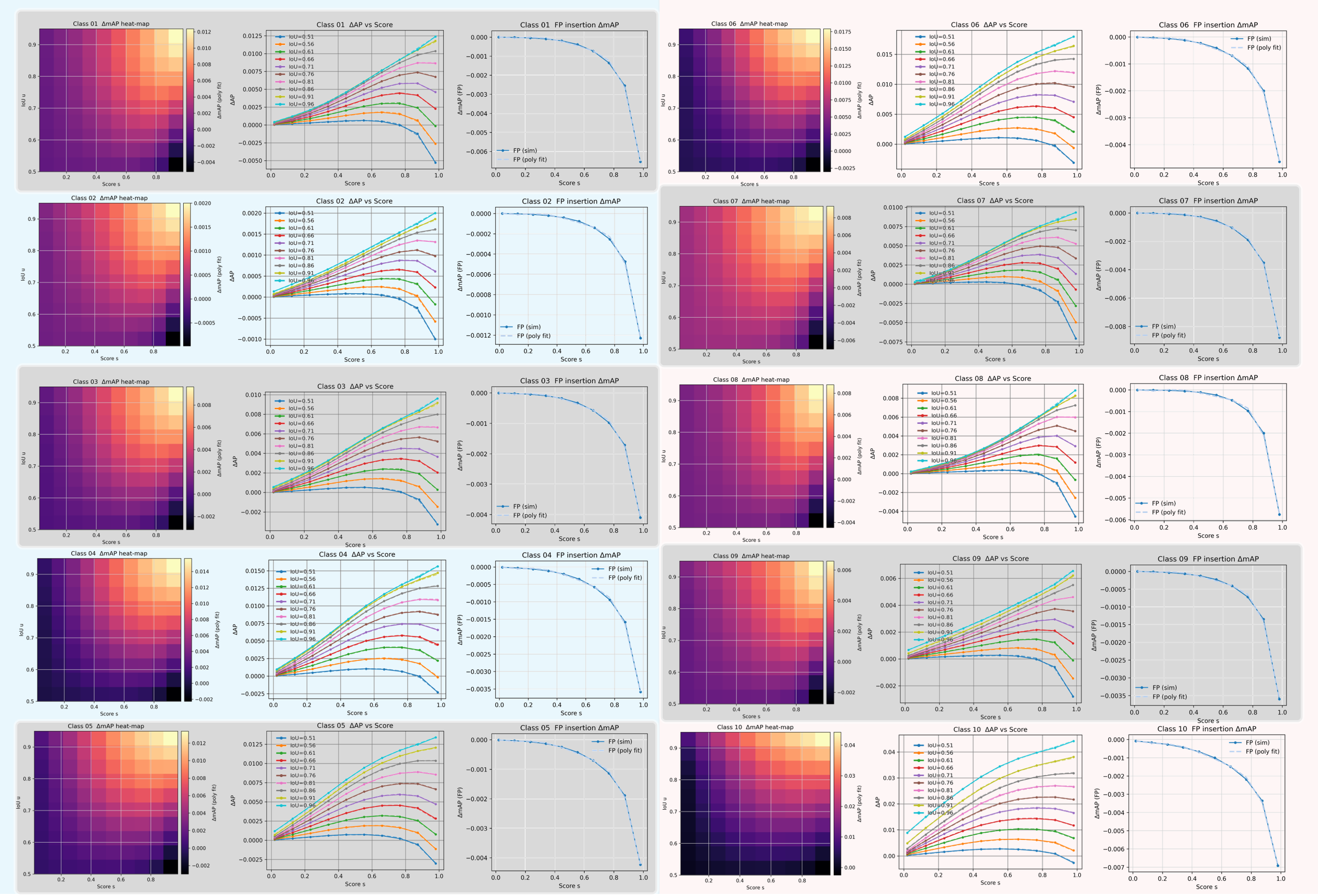}
\caption{\textbf{Comparison between class-wise $\Delta\mathrm{AP}$ functions under Faster R-CNN priors.}
Shown are the first ten COCO classes, where each column pair visualizes the analytic $\Delta\mathrm{AP}$ computed under the Beta-fitted prior. 
The 2D heatmaps (left) represent $\Delta\mathrm{AP}^{\mathrm{TP}}(s,\tau)$ with respect to score $s$ and IoU threshold $\tau$, while the 1D curves (right) show $\Delta\mathrm{AP}^{\mathrm{FP}}(s)$ for false-positive insertions. 
Both surfaces and curves are fitted with 6-order polynomials to enable efficient real-time computation of DetGain during online sampling.}
\label{fig:apdx-prior}
\end{figure*}

\paragraph{\textbf{Appendix 6. Ablation studies on prior distribution.}}
\begin{table}[t]
\centering
\footnotesize
\setlength{\tabcolsep}{5pt}
\renewcommand{\arraystretch}{1.2}
\caption{\textbf{Comparison between Faster R-CNN–specific and uniform priors.}
Results are obtained using a Faster R-CNN–Res50 student and a pretrained teacher of the same architecture.
The detector-specific prior is estimated from the teacher’s predictions on the COCO training set.}
\label{tab:prior-ablation}
\begin{tabular}{lcccc}
\toprule
\multirow{2}{*}{\textbf{Prior/Baseline}} 
& \multicolumn{2}{c}{\textbf{Train}} 
& \multicolumn{2}{c}{\textbf{Validation}} \\

\cmidrule(lr){2-3} \cmidrule(lr){4-5}
& \textbf{AP@50} & \textbf{AP@[50:95]} 
& \textbf{AP@50} & \textbf{AP@[50:95]} \\
\midrule
\rowcolor{rowgray}
Baseline & 67.4 & 44.5 & 58.3 & 37.4\\
Model–specific  & \textbf{68.8} & \textbf{45.6} & \textbf{61.2} & \textbf{39.6} \\
\rowcolor{rowgray}
Uniform Prior   & 67.6 & 45.1 & 60.7 & \textbf{39.6} \\
\bottomrule
\end{tabular}
\end{table}

We further compare the proposed \emph{uniform prior} with a detector-specific prior estimated from a pretrained Faster R-CNN (ResNet-50 backbone).  
The Faster R-CNN–specific distribution is obtained by maximum-likelihood estimation (MLE) over all predictions from the COCO \texttt{train2017} set.  
Since the model’s score distribution evolves during training and gradually aligns with the teacher’s output, we adopt the teacher’s prediction statistics as a stable approximation of the true detection prior for simulation.

Specifically, we collect all predictions across $80$ classes and $10$ IoU thresholds ($\mathrm{IoU}\in[0.5,0.95]$ with step $0.05$).  
For each $(c, \tau)$ pair, we record the empirical distributions of true positives (TP) and false positives (FP) and their total counts, then fit two Beta distributions via MLE.
In total, we obtain $80\times10=800$ Beta parameter pairs $(\alpha, \beta)$, each describing one class–IoU condition with its corresponding $T_{c,\tau}$ and $F_{c,\tau}$.
Using Eq.~\ref{eq:app-delta-ap-tp-final} and Eq.~\ref{eq:app-delta-ap-fp-final}, we compute the marginal change in AP ($\Delta\mathrm{AP}$) under these fitted priors through numerical integration, producing per-class $\Delta\mathrm{AP}$ surfaces with respect to IoU and score for TP insertion, and one-dimensional $\Delta\mathrm{AP}$ curves for FP insertion.
Fig.~\ref{fig:apdx-prior} visualizes the results for the first ten classes.
Both can be well approximated by high-order polynomial regression to replace numerical integration for real-time computation during online sampling.

By using the above procedure, we obtain the results in Table~\ref{tab:prior-ablation}.
While the model--specific prior is theoretically more accurate to the teacher's score distribution and indeed yields slightly higher AP on training set, its effect on \emph{validation} AP is negligible.
We conjecture two practical reasons: (i) the fitted prior captures teacher- and training-distribution idiosyncrasies that do not transfer to held-out data; and (ii) strong online augmentation already regularizes score distributions, narrowing the gap between priors.
Given the comparable convergence and final validation accuracy, together with its greater generality and zero fitting overhead, we adopt the \emph{uniform prior} for all detector families and use the unified closed-form in Eqs.~(\ref{eq:tp-closed-uniform}--\ref{eq:fp-closed-uniform}).

\paragraph{\textbf{Appendix 7. Experiments on additional datasets}}
We further evaluate the uniform-prior DetGain on datasets beyond COCO to assess robustness
under different data regimes and label distributions.
We report results on Pascal VOC 2007, 2012~\cite{voc} (small-scale generic detection) and BDD100K~\cite{bdd100k} (mid-scale driving with a long-tailed class distribution),
using Faster R-CNN with ResNet-50 and the standard $1\times$ training schedule.
We use the default training/validation splits: Pascal VOC has \textbf{16,551} training images and BDD100K has \textbf{69,863} training images and \textbf{10,000} validation images.

All other settings follow the default detection recipe of the corresponding codebase; DetGain is applied as the sample-selection criterion
within each super-batch, while the detector architecture, optimizer, and schedule are unchanged.As shown in Tab.~\ref{tab:extra_datasets}, DetGain consistently improves COCO-style mAP on both datasets:
on VOC2007 test (train set $\sim$16.5K), DetGain improves mAP from 51.3 to 54.3 (+3.0);
on BDD100K val (train set $\sim$70K), DetGain improves mAP from 30.3 to 32.1 (+1.8).
These gains support that the uniform-prior DetGain remains effective across dataset scales and distribution shifts.

\begin{table}[t]
  \centering
  \caption{\textbf{Uniform-prior DetGain on additional datasets.}
COCO-style mAP for Faster R-CNN-R50 on Pascal VOC2007 test and BDD100K validation using the default train/validation splits and the standard $1\times$ schedule.
DetGain consistently improves over the baseline (+3.0 on VOC2007, +1.8 on BDD100K). Single run for each dataset.}
  \label{tab:extra_datasets}
  \small
  \setlength{\tabcolsep}{7pt}
  \begin{tabular}{lcccc}
    \toprule
    Dataset  & Backbone & Baseline & +DetGain & $\Delta$ \\
    \midrule
    VOC2007  & R50 & 51.3 & 54.3 & +3.0 \\
    BDD100K  & R50 & 30.3 & 32.1 & +1.8 \\
    \bottomrule
  \end{tabular}
\end{table}

\begin{figure*}[h]
\centering
\includegraphics[width=0.8\linewidth]{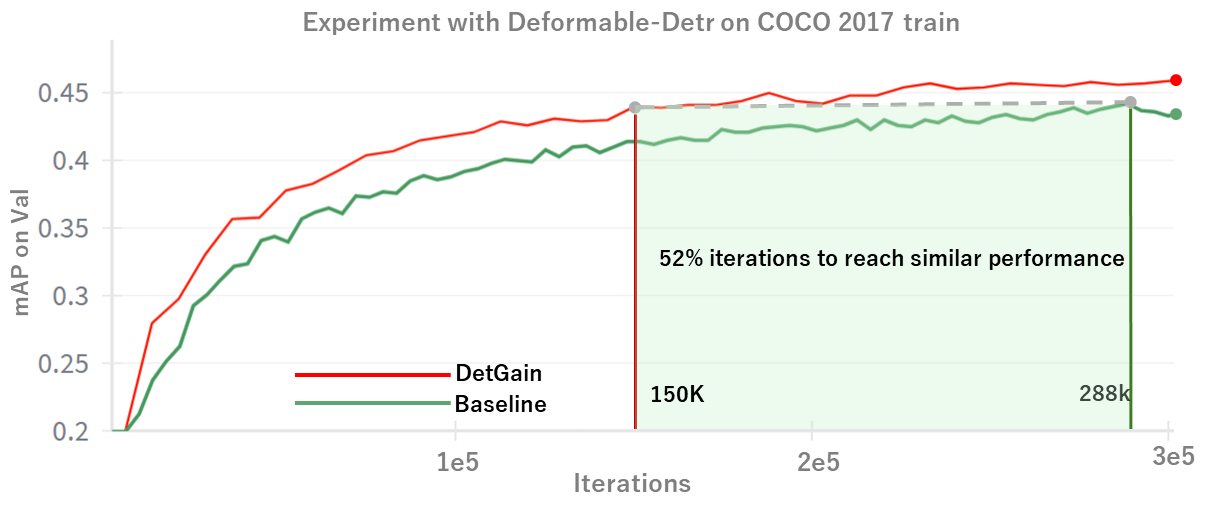}
\caption{\textbf{Performance curve with data curation.} We compare the performance curves of the baseline model and the model with data curation. 
The tested model is Deformable-DETR-Res50, trained with a batch size of 16 using the standard AdamW optimizer and a fixed learning rate of 1e-4. 
Data curation achieved the baseline’s best performance in nearly half the number of iterations.
}
\label{fig:apdx-prior}
\end{figure*}

\paragraph{\textbf{Appendix 8. Computational overhead and dynamic sampling.}}
DetGain adds a selection stage per iteration: we run an extra \texttt{no\_grad} + AMP scoring pass on the super-batch to compute per-image DetGain and select the top-$k$ images; the subsequent forward/backward update on the selected sub-batch is identical to the baseline.
Therefore, peak GPU memory is largely unchanged (dominated by the backward pass), while the wall-clock overhead is mainly determined by the super-batch scoring cost and the selection ratio $k$.

We report a specific case training Faster R-CNN-R50 on Pascal VOC, $k{=}20\%$, teacher = Faster R-CNN-R50.
Tab.~\ref{tab:detgain_overhead} reports iteration-level timing and memory.
With a naive implementation, the total iteration time increases from 0.25s to 0.75s.
Part of the increase comes from data loading / preprocessing due to reading the larger super-batch (0.0263s $\rightarrow$ 0.0726s) and strong data augmentation.
After subtracting this I/O difference, the remaining overhead is attributed to the scoring-and-selection stage.
GPU peak memory differs by less than 5\% between baseline and DetGain; CPU memory increases around 53\% due to buffering the super-batch.

\begin{table}[t]
\centering
\caption{\textbf{Runtime and memory overhead (Faster R-CNN-R50 on Pascal VOC, $k{=}20\%$).}
Per-iteration wall-clock time is decomposed into data loading/preprocessing and the remaining compute; the latter includes the DetGain scoring pass.
Percentages are relative to the baseline.}
\label{tab:detgain_overhead}
\footnotesize
\setlength{\tabcolsep}{6pt}
\begin{tabular}{lccc}
\toprule
Method & Iter time (s) & Data time (ms) & Non-data time (s)\\
\midrule
Baseline & 0.25 & 26.3 & 0.2237 \\
DetGain & 0.75 & 72.6 & 0.6774  \\
\midrule
Overhead (\%) & +200\% & +176\% & +203\%  \\
\bottomrule
\end{tabular}
\end{table}

\smallskip
\noindent\textbf{Effect of longer training schedules.}
DetGain's improvements are not a trivial artifact of extra compute.
On FCOS-R50, simply extending the baseline schedule yields diminishing returns and can even hurt validation performance due to overfitting:
the baseline saturates at $2\times$ (38.5 mAP) and drops at $3\times$ (37.1 mAP), whereas DetGain continues to improve monotonically from $0.5\times$ to $3\times$ (37.5 $\rightarrow$ 43.8 mAP; Tab.~\ref{tab:differentschedule}).
This suggests DetGain alters the optimization trajectory and mitigates late-stage overfitting, rather than merely benefiting from longer training.

To reduce overhead while preserving performance, we adopt a \emph{dynamic} selection ratio schedule:
we use a larger ratio early in training for more diverse representation learning, and a smaller ratio later for cost-efficient fine-tuning.
Concretely, we use $k{=}40\%$ for the first 60\% of iterations and $k{=}20\%$ for the remaining 40\%.
On FCOS-R50 with a $2\times$ schedule, this dynamic strategy reaches 42.5 mAP, retaining 98.8\% of the static-$20\%$ DetGain performance (42.5 vs.\ 43.0 mAP in Tab.~4), while reducing the total training time by approximately 30\% in practice.

\begin{table}[t]
\centering
\caption{\textbf{Effect of longer training schedules on FCOS-R50 (COCO val mAP).}
Baseline training saturates and overfits with longer schedules (performance drops at $3\times$),
while DetGain consistently improves across schedules.}
\label{tab:differentschedule}
\footnotesize
\setlength{\tabcolsep}{5pt}
\begin{tabular}{lcccc}
\toprule
Strategy & 0.5$\times$ & 1$\times$ & 2$\times$ & 3$\times$ \\
\midrule
Baseline & 33.4 & 38.2 & 38.5 & 37.1 (overfit)\\
DetGain &
37.5\,{\textcolor{gray}{(+4.1)}} &
40.9\,{\textcolor{gray}{(+2.7)}} &
43.0\,{\textcolor{gray}{(+4.5)}} &
43.8\,{\textcolor{gray}{(+6.7)}} \\
\bottomrule
\end{tabular}
\end{table}

\paragraph{Polynomial weight lookup.}
To avoid per-iteration density fitting while retaining a smooth dependence on confidence score and localization quality,
we pre-fit lightweight polynomial surrogates for the DetGain weights and evaluate them in $O(1)$ per prediction.
Specifically, we approximate the TP weight by a bivariate polynomial in IoU $u$ and score $s$,
and the FP weight by a univariate polynomial in score $s$:
\[
w_{\mathrm{TP}}(u,s,\ell)=\sum_{p+q\le n} c^{(\ell)}_{p,q}\,u^{p}s^{q},
w_{\mathrm{FP}}(s,\ell)=\sum_{r=0}^{m} d^{(\ell)}_{r}\,s^{r}.
\]
Coefficients $\{c^{(\ell)}_{p,q}\}$ and $\{d^{(\ell)}_{r}\}$ are fitted offline,
and optionally reweighted by a per-class scalar $\alpha_\ell$ to account for class frequency imbalance.
During scoring, we take the post-NMS top-$K$ predictions, match each prediction to its best-IoU ground truth,
apply the standard TP criterion $(u\ge\tau)$ with label agreement, and enforce the common detection convention that each ground-truth instance
contributes at most one TP for a given class by keeping only the highest-score TP among predictions assigned to the same $(\ell, \mathrm{GT})$ pair.
The final DetGain score of an image is the sum of TP and FP weights.

    \ifMain
    \else
        {
        \small
        \bibliographystyle{ieeenat_fullname}
        \bibliography{main}

@String(BMVC= {Brit. Mach. Vis. Conf.})

@String(BMVC  =	{BMVC})

@article{voc,
  author  = {Everingham, Mark and Van~Gool, Luc and Williams, Christopher K. I. and Winn, John and Zisserman, Andrew},
  title   = {The Pascal Visual Object Classes (VOC) Challenge},
  journal = {International Journal of Computer Vision},
  volume  = {88},
  number  = {2},
  pages   = {303--338},
  month   = {June},
  year    = {2010}
}

@inproceedings{bdd100k,
  title={Bdd100k: A diverse driving dataset for heterogeneous multitask learning},
  author={Yu, Fisher and Chen, Haofeng and Wang, Xin and Xian, Wenqi and Chen, Yingying and Liu, Fangchen and Madhavan, Vashisht and Darrell, Trevor},
  booktitle={Proceedings of the IEEE/CVF conference on computer vision and pattern recognition},
  pages={2636--2645},
  year={2020}
}

@article{loshchilov2015online,
  title={Online batch selection for faster training of neural networks},
  author={Loshchilov, Ilya and Hutter, Frank},
  journal={arXiv preprint arXiv:1511.06343},
  year={2015}
}

@inproceedings{kawaguchi2020ordered,
  title={Ordered sgd: A new stochastic optimization framework for empirical risk minimization},
  author={Kawaguchi, Kenji and Lu, Haihao},
  booktitle={International Conference on Artificial Intelligence and Statistics},
  pages={669--679},
  year={2020},
  organization={PMLR}
}

@inproceedings{jiang2018mentornet,
  title={Mentornet: Learning data-driven curriculum for very deep neural networks on corrupted labels},
  author={Jiang, Lu and Zhou, Zhengyuan and Leung, Thomas and Li, Li-Jia and Fei-Fei, Li},
  booktitle={International conference on machine learning},
  pages={2304--2313},
  year={2018},
  organization={PMLR}
}

@article{li2006confidence,
  title={Confidence-based active learning},
  author={Li, Mingkun and Sethi, Ishwar K},
  journal={IEEE transactions on pattern analysis and machine intelligence},
  volume={28},
  number={8},
  pages={1251--1261},
  year={2006},
  publisher={IEEE}
}

@article{coleman2019selection,
  title={Selection via proxy: Efficient data selection for deep learning},
  author={Coleman, Cody and Yeh, Christopher and Mussmann, Stephen and Mirzasoleiman, Baharan and Bailis, Peter and Liang, Percy and Leskovec, Jure and Zaharia, Matei},
  journal={arXiv preprint arXiv:1906.11829},
  year={2019}
}

@inproceedings{killamsetty2021grad,
  title={Grad-match: Gradient matching based data subset selection for efficient deep model training},
  author={Killamsetty, Krishnateja and Durga, Sivasubramanian and Ramakrishnan, Ganesh and De, Abir and Iyer, Rishabh},
  booktitle={International Conference on Machine Learning},
  pages={5464--5474},
  year={2021},
  organization={PMLR}
}

@article{toneva2018empirical,
  title={An empirical study of example forgetting during deep neural network learning},
  author={Toneva, Mariya and Sordoni, Alessandro and Combes, Remi Tachet des and Trischler, Adam and Bengio, Yoshua and Gordon, Geoffrey J},
  journal={arXiv preprint arXiv:1812.05159},
  year={2018}
}

@article{johnson2018training,
  title={Training deep models faster with robust, approximate importance sampling},
  author={Johnson, Tyler B and Guestrin, Carlos},
  journal={Advances in Neural Information Processing Systems},
  volume={31},
  year={2018}
}

@article{han2018co,
  title={Co-teaching: Robust training of deep neural networks with extremely noisy labels},
  author={Han, Bo and Yao, Quanming and Yu, Xingrui and Niu, Gang and Xu, Miao and Hu, Weihua and Tsang, Ivor and Sugiyama, Masashi},
  journal={Advances in neural information processing systems},
  volume={31},
  year={2018}
}

@article{hao2025progressive,
  title={Progressive Data Dropout: An Embarrassingly Simple Approach to Faster Training},
  author={Hao, Xinyue and Hou, Shihao and Lu, Yang and Sevilla-Lara, Laura and Arnab, Anurag and Gowda, Shreyank N and others},
  journal={arXiv preprint arXiv:2505.22342},
  year={2025}
}

@inproceedings{rholoss,
  title={Prioritized training on points that are learnable, worth learning, and not yet learnt},
  author={Mindermann, S{\"o}ren and Brauner, Jan M and Razzak, Muhammed T and Sharma, Mrinank and Kirsch, Andreas and Xu, Winnie and H{\"o}ltgen, Benedikt and Gomez, Aidan N and Morisot, Adrien and Farquhar, Sebastian and others},
  booktitle={International Conference on Machine Learning},
  pages={15630--15649},
  year={2022},
  organization={PMLR}
}

@inproceedings{jest,
 author = {Evans, Talfan and Parthasarathy, Nikhil and Merzi\'{c}, Hamza and H\'{e}naff, Olivier J.},
 booktitle = {Advances in Neural Information Processing Systems},

 pages = {141240--141260},
 publisher = {Curran Associates, Inc.},
 title = {Data curation via joint example selection further accelerates multimodal learning},
 
 volume = {37},
 year = {2024}
}

@inproceedings{aced,
  title={Active data curation effectively distills large-scale multimodal models},
  author={Udandarao, Vishaal and Parthasarathy, Nikhil and Naeem, Muhammad Ferjad and Evans, Talfan and Albanie, Samuel and Tombari, Federico and Xian, Yongqin and Tonioni, Alessio and H{\'e}naff, Olivier J},
  booktitle={Proceedings of the Computer Vision and Pattern Recognition Conference},
  pages={14422--14437},
  year={2025}
}

@article{liu2020deep,
  title={Deep learning for generic object detection: A survey},
  author={Liu, Li and Ouyang, Wanli and Wang, Xiaogang and Fieguth, Paul and Chen, Jie and Liu, Xinwang and Pietik{\"a}inen, Matti},
  journal={International journal of computer vision},
  volume={128},
  number={2},
  pages={261--318},
  year={2020},
  publisher={Springer}
}

@inproceedings{cao2020prime,
  title={Prime sample attention in object detection},
  author={Cao, Yuhang and Chen, Kai and Loy, Chen Change and Lin, Dahua},
  booktitle={Proceedings of the IEEE/CVF conference on computer vision and pattern recognition},
  pages={11583--11591},
  year={2020}
}

@inproceedings{choi2021active,
  title={Active learning for deep object detection via probabilistic modeling},
  author={Choi, Jiwoong and Elezi, Ismail and Lee, Hyuk-Jae and Farabet, Clement and Alvarez, Jose M},
  booktitle={Proceedings of the IEEE/CVF international conference on computer vision},
  pages={10264--10273},
  year={2021}
}

@article{wang2024data,
  title={Data Shapley in One Training Run},
  author={Wang, Jiachen T and Mittal, Prateek and Song, Dawn and Jia, Ruoxi},
  journal={arXiv preprint arXiv:2406.11011},
  year={2024}
  }

@inproceedings{mirzasoleiman2020coresets,
  title={Coresets for data-efficient training of machine learning models},
  author={Mirzasoleiman, Baharan and Bilmes, Jeff and Leskovec, Jure},
  booktitle={International Conference on Machine Learning},
  pages={6950--6960},
  year={2020},
  organization={PMLR}
}

@article{superloss,
  title={Superloss: A generic loss for robust curriculum learning},
  author={Castells, Thibault and Weinzaepfel, Philippe and Revaud, Jerome},
  journal={Advances in Neural Information Processing Systems},
  volume={33},
  pages={4308--4319},
  year={2020}
}

@inproceedings{shah2020choosing,
  title={Choosing the sample with lowest loss makes sgd robust},
  author={Shah, Vatsal and Wu, Xiaoxia and Sanghavi, Sujay},
  booktitle={International Conference on Artificial Intelligence and Statistics},
  pages={2120--2130},
  year={2020},
  organization={PMLR}
}

@inproceedings{roy2018deep,
  title={Deep active learning for object detection.},
  author={Roy, Soumya and Unmesh, Asim and Namboodiri, Vinay P},
  booktitle={BMVC},
  volume={362},
  number={91},
  pages={375},
  year={2018}
}

@inproceedings{kao2018localization,
  title={Localization-aware active learning for object detection},
  author={Kao, Chieh-Chi and Lee, Teng-Yok and Sen, Pradeep and Liu, Ming-Yu},
  booktitle={Asian Conference on Computer Vision},
  pages={506--522},
  year={2018},
  organization={Springer}
}

@inproceedings{agarwal2020contextual,
  title={Contextual diversity for active learning},
  author={Agarwal, Sharat and Arora, Himanshu and Anand, Saket and Arora, Chetan},
  booktitle={European Conference on Computer Vision},
  pages={137--153},
  year={2020},
  organization={Springer}
}

@inproceedings{yuan2021multiple,
  title={Multiple instance active learning for object detection},
  author={Yuan, Tianning and Wan, Fang and Fu, Mengying and Liu, Jianzhuang and Xu, Songcen and Ji, Xiangyang and Ye, Qixiang},
  booktitle={Proceedings of the IEEE/CVF Conference on Computer Vision and Pattern Recognition},
  pages={5330--5339},
  year={2021}
}

@inproceedings{wu2022entropy,
  title={Entropy-based active learning for object detection with progressive diversity constraint},
  author={Wu, Jiaxi and Chen, Jiaxin and Huang, Di},
  booktitle={Proceedings of the IEEE/CVF conference on computer vision and pattern recognition},
  pages={9397--9406},
  year={2022}
}

@inproceedings{yang2024plug,
  title={Plug and play active learning for object detection},
  author={Yang, Chenhongyi and Huang, Lichao and Crowley, Elliot J},
  booktitle={Proceedings of the IEEE/CVF conference on computer vision and pattern recognition},
  pages={17784--17793},
  year={2024}
}

@inproceedings{mi2022active,
  title={Active teacher for semi-supervised object detection},
  author={Mi, Peng and Lin, Jianghang and Zhou, Yiyi and Shen, Yunhang and Luo, Gen and Sun, Xiaoshuai and Cao, Liujuan and Fu, Rongrong and Xu, Qiang and Ji, Rongrong},
  booktitle={Proceedings of the IEEE/CVF conference on computer vision and pattern recognition},
  pages={14482--14491},
  year={2022}
}

@article{sener2017active,
  title={Active learning for convolutional neural networks: A core-set approach},
  author={Sener, Ozan and Savarese, Silvio},
  journal={arXiv preprint arXiv:1708.00489},
  year={2017}
}

@article{borsos2020coresets,
  title={Coresets via bilevel optimization for continual learning and streaming},
  author={Borsos, Zal{\'a}n and Mutny, Mojmir and Krause, Andreas},
  journal={Advances in neural information processing systems},
  volume={33},
  pages={14879--14890},
  year={2020}
}

@inproceedings{lee2024coreset,
  title={Coreset selection for object detection},
  author={Lee, Hojun and Kim, Suyoung and Lee, Junhoo and Yoo, Jaeyoung and Kwak, Nojun},
  booktitle={Proceedings of the IEEE/CVF Conference on Computer Vision and Pattern Recognition},
  pages={7682--7691},
  year={2024}
}

@inproceedings{zhou2024optimizing,
  title={Optimizing object detection via metric-driven training data selection},
  author={Zhou, Changyuan and Guo, Yumin and Lv, Qinxue and Yuan, Ji},
  booktitle={Proceedings of the IEEE/CVF Conference on Computer Vision and Pattern Recognition},
  pages={7348--7355},
  year={2024}
}

@article{oquab2023dinov2,
  title={Dinov2: Learning robust visual features without supervision},
  author={Oquab, Maxime and Darcet, Timoth{\'e}e and Moutakanni, Th{\'e}o and Vo, Huy and Szafraniec, Marc and Khalidov, Vasil and Fernandez, Pierre and Haziza, Daniel and Massa, Francisco and El-Nouby, Alaaeldin and others},
  journal={arXiv preprint arXiv:2304.07193},
  year={2023}
}

@article{gunasekar2023textbooks,
  title={Textbooks are all you need},
  author={Gunasekar, Suriya and Zhang, Yi and Aneja, Jyoti and Mendes, Caio C{\'e}sar Teodoro and Del Giorno, Allie and Gopi, Sivakanth and Javaheripi, Mojan and Kauffmann, Piero and de Rosa, Gustavo and Saarikivi, Olli and others},
  journal={arXiv preprint arXiv:2306.11644},
  year={2023}
}

@article{liu2024deepseek,
  title={Deepseek-v3 technical report},
  author={Liu, Aixin and Feng, Bei and Xue, Bing and Wang, Bingxuan and Wu, Bochao and Lu, Chengda and Zhao, Chenggang and Deng, Chengqi and Zhang, Chenyu and Ruan, Chong and others},
  journal={arXiv preprint arXiv:2412.19437},
  year={2024}
}

@article{abbas2023semdedup,
  title={Semdedup: Data-efficient learning at web-scale through semantic deduplication},
  author={Abbas, Amro and Tirumala, Kushal and Simig, D{\'a}niel and Ganguli, Surya and Morcos, Ari S},
  journal={arXiv preprint arXiv:2303.09540},
  year={2023}
}

@inproceedings{evans2024bad,
  title={Bad students make great teachers: Active learning accelerates large-scale visual understanding},
  author={Evans, Talfan and Pathak, Shreya and Merzic, Hamza and Schwarz, Jonathan and Tanno, Ryutaro and Henaff, Olivier J},
  booktitle={European Conference on Computer Vision},
  pages={264--280},
  year={2024},
  organization={Springer}
}

@inproceedings{vfnet,
  title={Varifocalnet: An iou-aware dense object detector},
  author={Zhang, Haoyang and Wang, Ying and Dayoub, Feras and Sunderhauf, Niko},
  booktitle={Proceedings of the IEEE/CVF conference on computer vision and pattern recognition},
  pages={8514--8523},
  year={2021}
}

@article{fasterrcnn,
  title={Faster R-CNN: Towards real-time object detection with region proposal networks},
  author={Ren, Shaoqing and He, Kaiming and Girshick, Ross and Sun, Jian},
  journal={IEEE transactions on pattern analysis and machine intelligence},
  volume={39},
  number={6},
  pages={1137--1149},
  year={2016},
  publisher={IEEE}
}

@inproceedings{atss,
  title={Bridging the gap between anchor-based and anchor-free detection via adaptive training sample selection},
  author={Zhang, Shifeng and Chi, Cheng and Yao, Yongqiang and Lei, Zhen and Li, Stan Z},
  booktitle={Proceedings of the IEEE/CVF conference on computer vision and pattern recognition},
  pages={9759--9768},
  year={2020}
}

@inproceedings{detr,
  title={End-to-end object detection with transformers},
  author={Carion, Nicolas and Massa, Francisco and Synnaeve, Gabriel and Usunier, Nicolas and Kirillov, Alexander and Zagoruyko, Sergey},
  booktitle={European conference on computer vision},
  pages={213--229},
  year={2020},
  organization={Springer}
}

@inproceedings{dino,
  title={DINO: DETR with Improved DeNoising Anchor Boxes for End-to-End Object Detection},
  author={Zhang, Hao and Li, Feng and Liu, Shilong and Zhang, Lei and Su, Hang and Zhu, Jun and Ni, Lionel and Shum, Heung-Yeung},
  year={2023},
  booktitle={The Eleventh International Conference on Learning Representations}
}

@inproceedings{deformable,
  title={Deformable DETR: Deformable Transformers for End-to-End Object Detection},
  author={Zhu, Xizhou and Su, Weijie and Lu, Lewei and Li, Bin and Wang, Xiaogang and Dai, Jifeng},
  year = {2021},
  booktitle={International Conference on Learning Representations}
}

@inproceedings{fcos,
  title={Fcos: Fully convolutional one-stage object detection},
  author={Tian, Zhi and Shen, Chunhua and Chen, Hao and He, Tong},
  booktitle={Proceedings of the IEEE/CVF international conference on computer vision},
  pages={9627--9636},
  year={2019}
}

@inproceedings{retinanet,
  title={Focal loss for dense object detection},
  author={Lin, Tsung-Yi and Goyal, Priya and Girshick, Ross and He, Kaiming and Doll{\'a}r, Piotr},
  booktitle={Proceedings of the IEEE international conference on computer vision},
  pages={2980--2988},
  year={2017}
}

@inproceedings{COCO2017,
  title={Microsoft coco: Common objects in context},
  author={Lin, Tsung-Yi and Maire, Michael and Belongie, Serge and Hays, James and Perona, Pietro and Ramanan, Deva and Doll{\'a}r, Piotr and Zitnick, C Lawrence},
  booktitle={European conference on computer vision},
  pages={740--755},
  year={2014},
  organization={Springer}
}

@inproceedings{ssd,
  title={Ssd: Single shot multibox detector},
  author={Liu, Wei and Anguelov, Dragomir and Erhan, Dumitru and Szegedy, Christian and Reed, Scott and Fu, Cheng-Yang and Berg, Alexander C},
  booktitle={European conference on computer vision},
  pages={21--37},
  year={2016},
  organization={Springer}
}

@inproceedings{ohem,
  title={Training region-based object detectors with online hard example mining},
  author={Shrivastava, Abhinav and Gupta, Abhinav and Girshick, Ross},
  booktitle={Proceedings of the IEEE conference on computer vision and pattern recognition},
  pages={761--769},
  year={2016}
}

@inproceedings{crosskd,
  title={CrossKD: Cross-head knowledge distillation for object detection},
  author={Wang, Jiabao and Chen, Yuming and Zheng, Zhaohui and Li, Xiang and Cheng, Ming-Ming and Hou, Qibin},
  booktitle={Proceedings of the IEEE/CVF conference on computer vision and pattern recognition},
  pages={16520--16530},
  year={2024}
}

@article{pkd,
  title={Pkd: General distillation framework for object detectors via pearson correlation coefficient},
  author={Cao, Weihan and Zhang, Yifan and Gao, Jianfei and Cheng, Anda and Cheng, Ke and Cheng, Jian},
  journal={Advances in Neural Information Processing Systems},
  volume={35},
  pages={15394--15406},
  year={2022}
}

@inproceedings{aploss1,
  title={Towards accurate one-stage object detection with ap-loss},
  author={Chen, Kean and Li, Jianguo and Lin, Weiyao and See, John and Wang, Ji and Duan, Lingyu and Chen, Zhibo and He, Changwei and Zou, Junni},
  booktitle={Proceedings of the IEEE/CVF Conference on Computer Vision and Pattern Recognition},
  pages={5119--5127},
  year={2019}
}

@article{aploss2,
  title={Robust and decomposable average precision for image retrieval},
  author={Ramzi, Elias and Thome, Nicolas and Rambour, Cl{\'e}ment and Audebert, Nicolas and Bitot, Xavier},
  journal={Advances in Neural Information Processing Systems},
  volume={34},
  pages={23569--23581},
  year={2021}
}

@article{mmdetection,
  title   = {{MMDetection}: Open MMLab Detection Toolbox and Benchmark},
  author  = {Chen, Kai and Wang, Jiaqi and Pang, Jiangmiao and Cao, Yuhang and
             Xiong, Yu and Li, Xiaoxiao and Sun, Shuyang and Feng, Wansen and
             Liu, Ziwei and Xu, Jiarui and Zhang, Zheng and Cheng, Dazhi and
             Zhu, Chenchen and Cheng, Tianheng and Zhao, Qijie and Li, Buyu and
             Lu, Xin and Zhu, Rui and Wu, Yue and Dai, Jifeng and Wang, Jingdong
             and Shi, Jianping and Ouyang, Wanli and Loy, Chen Change and Lin, Dahua},
  journal= {arXiv preprint arXiv:1906.07155},
  year={2019}
}

@article{gfl,
  title={Generalized focal loss: Learning qualified and distributed bounding boxes for dense object detection},
  author={Li, Xiang and Wang, Wenhai and Wu, Lijun and Chen, Shuo and Hu, Xiaolin and Li, Jun and Tang, Jinhui and Yang, Jian},
  journal={Advances in neural information processing systems},
  volume={33},
  pages={21002--21012},
  year={2020}
}

@article{adamw,
  title={Decoupled weight decay regularization},
  author={Loshchilov, Ilya and Hutter, Frank},
  journal={arXiv preprint arXiv:1711.05101},
  year={2017}
}

@inproceedings{resnet,
  title={Deep residual learning for image recognition},
  author={He, Kaiming and Zhang, Xiangyu and Ren, Shaoqing and Sun, Jian},
  booktitle={Proceedings of the IEEE conference on computer vision and pattern recognition},
  pages={770--778},
  year={2016}
}

@inproceedings{resnetxt,
  title={Aggregated residual transformations for deep neural networks},
  author={Xie, Saining and Girshick, Ross and Doll{\'a}r, Piotr and Tu, Zhuowen and He, Kaiming},
  booktitle={Proceedings of the IEEE conference on computer vision and pattern recognition},
  pages={1492--1500},
  year={2017}
}

@inproceedings{swin,
  title={Swin transformer: Hierarchical vision transformer using shifted windows},
  author={Liu, Ze and Lin, Yutong and Cao, Yue and Hu, Han and Wei, Yixuan and Zhang, Zheng and Lin, Stephen and Guo, Baining},
  booktitle={Proceedings of the IEEE/CVF international conference on computer vision},
  pages={10012--10022},
  year={2021}
}

@inproceedings{yolov8,
  title={Yolov8: A novel object detection algorithm with enhanced performance and robustness},
  author={Varghese, Rejin and Sambath, M},
  booktitle={2024 International conference on advances in data engineering and intelligent computing systems (ADICS)},
  pages={1--6},
  year={2024},
  organization={IEEE}
}

@Article{albu,
    AUTHOR = {Buslaev, Alexander and Iglovikov, Vladimir I. and Khvedchenya, Eugene and Parinov, Alex and Druzhinin, Mikhail and Kalinin, Alexandr A.},
    TITLE = {Albumentations: Fast and Flexible Image Augmentations},
    JOURNAL = {Information},
    VOLUME = {11},
    YEAR = {2020},
    NUMBER = {2},
    ARTICLE-NUMBER = {125},
    URL = {https://www.mdpi.com/2078-2489/11/2/125},
    ISSN = {2078-2489},
    DOI = {10.3390/info11020125}
}

@inproceedings{lkd,
  title={Localization distillation for dense object detection},
  author={Zheng, Zhaohui and Ye, Rongguang and Wang, Ping and Ren, Dongwei and Zuo, Wangmeng and Hou, Qibin and Cheng, Ming-Ming},
  booktitle={Proceedings of the IEEE/CVF conference on computer vision and pattern recognition},
  pages={9407--9416},
  year={2022}
}

@inproceedings{mulkd,
  title={Learning lightweight object detectors via multi-teacher progressive distillation},
  author={Cao, Shengcao and Li, Mengtian and Hays, James and Ramanan, Deva and Wang, Yu-Xiong and Gui, Liangyan},
  booktitle={International Conference on Machine Learning},
  pages={3577--3598},
  year={2023},
  organization={PMLR}
}
        }
    \fi
\fi

\end{document}